\pdfoutput=1
\documentclass[11pt]{article}

\usepackage{acl}
\usepackage{times}
\usepackage{latexsym}
\usepackage{enumitem}
\usepackage[T1]{fontenc}
\usepackage[utf8]{inputenc}

\usepackage{microtype}

\usepackage{inconsolata}
\usepackage{booktabs}
\usepackage{booktabs}
\usepackage{multirow}
\usepackage{amsmath}

\usepackage[ruled, vlined, linesnumbered, commentsnumbered, longend]{algorithm2e}

\usepackage{xcolor}

\usepackage{graphics}
\usepackage{graphicx}
\usepackage{subfig}
\usepackage[capitalize]{cleveref}

\usepackage[english]{babel}

\usepackage{amsfonts}

\usepackage{amssymb} 
\usepackage{marvosym}
\usepackage{graphicx}
\usepackage{subcaption}
\usepackage[capitalize]{cleveref}

\title{\emph{How Many Validation Labels Do You Need?} \\ Exploring the Design Space of Label-Efficient Model Ranking}

\author{Zhengyu Hu$^1$ \quad Jieyu Zhang$^2$ \quad Yue Yu$^3$  \quad Yuchen Zhuang$^3$   \quad Hui Xiong$^1$\Letter \\ 
$^1$ HKUST (GZ)  \quad  $^2$ University of Washington  \quad $^3$ Georgia Institute of Technology \\
\texttt{zhu021@connect.hkust-gz.edu.cn},  \texttt{jieyuz2@cs.washington.edu},\\
\texttt{\{yueyu, yczhuang\}@gatech.edu}, \texttt{xionghui@ust.hk}\\ 
  }

\usepackage{xspace}

\def \OURS {LEMR}

\usepackage{xspace}
\makeatletter
\DeclareRobustCommand\onedot{\futurelet\@let@token\@onedot}
\def\@onedot{\ifx\@let@token.\else.\null\fi\xspace}

\begin{document}

\newcommand\mycommfont[1]{\small\ttfamily\textcolor{blue}{#1}}
\SetCommentSty{mycommfont}

\maketitle
\begin{abstract}

This paper presents \OURS{} (Label-Efficient Model Ranking) and introduces the MoraBench Benchmark. 
\OURS{} is a novel framework that minimizes the need for costly annotations in model selection by strategically annotating instances from an unlabeled validation set.
To evaluate \OURS{}, we leverage the MoraBench Benchmark, a comprehensive collection of model outputs across diverse scenarios.
Our extensive evaluation across 23 different NLP tasks in semi-supervised learning, weak supervision, and prompt selection tasks demonstrates \OURS{}'s effectiveness in significantly reducing labeling costs. 
Key findings highlight the impact of suitable ensemble methods, uncertainty sampling strategies, and model committee selection in enhancing model ranking accuracy. 
\OURS{}, supported by the insights from MoraBench, provides a cost-effective and accurate solution for model selection, especially valuable in resource-constrained environments.

\end{abstract}
\section{Introduction}
Model selection plays a central role in building robust predictive systems for Natural Language Processing (NLP)~\cite{awasthy2020predictive, Lizotte2021ModelS, zhang2022targeted, han2023fair}, which underpins numerous application scenarios including feature engineering~\cite{severyn2013automatic}, algorithm selection~\cite{Yang2023TestAV}, and hyperparameter tuning~\cite{liu-wang-2021-empirical}.
Typically, in a standard machine learning pipeline, a held-out validation set is utilized for the model selection purpose, which often contains massive labeled data. 
Under a more practical low-resource setting, however, creating a large set of validation data is no longer feasible~\cite{perez2021true,bragg2021flex} due to the additional annotation cost~\citep{zhang2023hypertime} as well as the reliance on domain expertise~\citep{hu2023leveraging}.
The resolution of this challenge is vital for the deployment of model selection techniques under real application scenarios.

Facilitating model selection under the true resource-limited scenarios can be challenging. 
Existing approaches often adopt fixed parameter~\citep{liu2022fewshot}, or early stopping~\citep{mahsereci2017early,choi2022early} for model selection, yet it can suffer from the training instability issue under the low-resource settings and does not reliably choose better-than-average hyperparameters~\citep{blier2018description,perez2021true}. 
There are also several works~\citep{zhou2022prompt,lu2022} that focus on \emph{unsupervised model selection}, which creates pseudo-validation sets for ranking different models. 
Nevertheless, without labeled data, there often exists a significant disparity between the ranking results produced by these methods and the true model rankings.
In summary,  model ranking remains challenging and under-explored under low-resource scenarios.

In this work, we propose \OURS{}  (\textbf{L}abel-\textbf{E}fficient \textbf{M}odel \textbf{R}anking), a framework that significantly reduces the need for costly annotations.
Our framework operates without presuming the availability of ground-truth clean labels.
Instead, we aim to \emph{strategically annotate} instances from an unlabeled validation set for model ranking. 
The framework can be divided into four steps.
First, an ensemble method with a selected model committee generates pseudo-labels for examples from the validation set, reducing the labeling cost  
(\textbf{Step-I} in Section~\ref{sec:plg}). 
Subsequently, we address the inherent noise in these pseudo-labels through two strategies: 
We first use uncertainty sampling to acquire ground-truth labels (\textbf{Step-II} in Section~\ref{sec:tlr}).,  and then utilize a Z-score mechanism to dynamically adjust the model committee based on these updated labels, further refining the labeling process (\textbf{Step-III} in Section~\ref{sec:msu}).
Finally, \OURS{} ranks all models using the refined pseudo-label and ground-truth label sets (\textbf{Step-IV} in Section~\ref{sec:mr}).  
This framework allows us to create a design space for model ranking, facilitating a systematic exploration of the efficacy across different selection metrics and identifying optimal strategies for each stage.

Specifically, we first organize the intersection for our framework \OURS{} by proposing an explicit design space centered around disentangling the following key methodological considerations:

\begin{itemize}[leftmargin=0.4cm]
\item \textit{Pseudo labels generation  (Section~\ref{sec:plg}): How to generate pseudo-labels?} 
We adopt an ensemble method based on our model committee to obtain the pseudo-labels. 
Two variants, soft ensemble, and hard ensemble~\cite{krogh1994neural,hansen1990neural}, are considered for this purpose.

\item \textit{Label Acquiring  (Section~\ref{sec:tlr}): Which of the pseudo-labels needs to be acquired?} 
Given the presence of noise in pseudo-labels, acquiring ground-truth labels is sometimes necessary. 
We employ uncertainty sampling strategies to identify which pseudo-labels to replace. 
Our approach includes uncertainty, classification margin, entropy, and random sampling strategies.

\item \textit{Model Committee Selection  (Section~\ref{sec:msu}): How to select a model committee reasonably?} 
Selecting an appropriate model committee is crucial.
We propose two methods: Z-score and All-model. 
The choice between them depends on balancing the desire for precision (favoring the Z-score method) and the need for diversity and comprehensive coverage (favoring the All-model approach).

\end{itemize}

With our design space, we can organize different methods and modularly generate a variety of methods. 
To evaluate these methods and facilitate future research in model ranking, we introduce the MoraBench (\textbf{Mo}del \textbf{Ra}nking \textbf{Bench}mark) in Section~\ref{sec:MoraBench}. 
It covers diverse scenarios, including semi-supervised learning  (Section~\ref{exp:semi}), weak supervision  (Section~\ref{exp:weak}), and prompt selection  (Section~\ref{exp:prompt}) tasks with 23 different tasks.
The experiments on MoraBench lead to the following observations:

\begin{itemize}[leftmargin=0.4cm]
\item 
With a suitable combination of methods within the design space, our framework can dramatically reduce the labeling cost for model selection.
For instance, in the semi-supervised learning scenario (AGNews task), labeling just 387 samples suffices for model selection, compared to the conventional need for 2000 samples.

\item In Pseudo-label Generation Step (Section~\ref{sec:plg}), under a limited budget, we find that soft ensemble yields a higher quality model ranking if the model in the model set performs poorly, otherwise hard ensemble is a better choice.

\item In Active Label Acquisition Step (Section~\ref{sec:tlr}), our findings underline the superiority of uncertainty sampling over random acquisition in all tasks.

\item In Model Committee Selection Step (Section~\ref{sec:msu}), 
We observe that a high-quality committee crucially influences the quality of model ranking. 
For this reason, a  Z-score-based selection method is designed, which outperforms the All-model strategy on all datasets.

\end{itemize}

\section{Related Work}

\subsection{Pseudo-labeling}
Lately, pseudo-labeling has marked a significant progression in deep learning, utilizing models to predict unlabeled data samples~\cite{lee2013pseudo,chen-etal-2021-revisiting,xu2023neighborhood,yang-etal-2023-prototype,zhang2021creating}. 
\citet{zhu2023towards} explore self-adaptive pseudo-label filtering, aiming to refine the selection process for pseudo-labels to boost learning performance.
Another popular technique is ensemble distillation~\cite{bachman2014learning,hinton2015distilling,hu2023leveraging}, which means distilling knowledge in an ensemble into a single model.

\subsection{Model Selection}
Model selection~\cite{Kohavi1995ASO,Kayali2022MiningRD,zhang2023hypertime} refers to determining the best from a set of candidate models based on their performance on a given dataset.
In the domain of this area, current research encompasses a variety of innovative methodologies, especially in the field of natural language processing~\cite{Yang2023TestAV,han2023fair,Du2021AllNT}.
\citet{lu2022} leverage the entropy statistics to select the best prompt orders for in-context learning. 
\citet{zhou2022prompt} propose an unsupervised model selection criterion that encourages consistency but simultaneously penalizes collapse.

\section{Preliminaries}

In this work, we consider a $C$-way classification task $\mathcal{T}$. 
For task $\mathcal{T}$, there exists $K$ trained models, denoted as $\mathcal{M} = \{ m_{k} \}_{k \in [K]}$. Our objective is to rank these models so that top-ranked models will achieve better performance on $\mathcal{T}$. Importantly, we work under the constraint of having no access to the original training data, instead relying on an unlabeled validation set $D_{V} =  \{x_{i}\}_{i \in [N]}$, along with a limited annotation budget $B$.

Our primary goal is to optimize the annotation process for the validation set in the context of model selection. 
To this end, we systematically study the effectiveness of our framework across different selection metrics and determine the optimal methods and timing for its utilization.

\begin{figure*}
    \centering
    \includegraphics[scale=0.5]{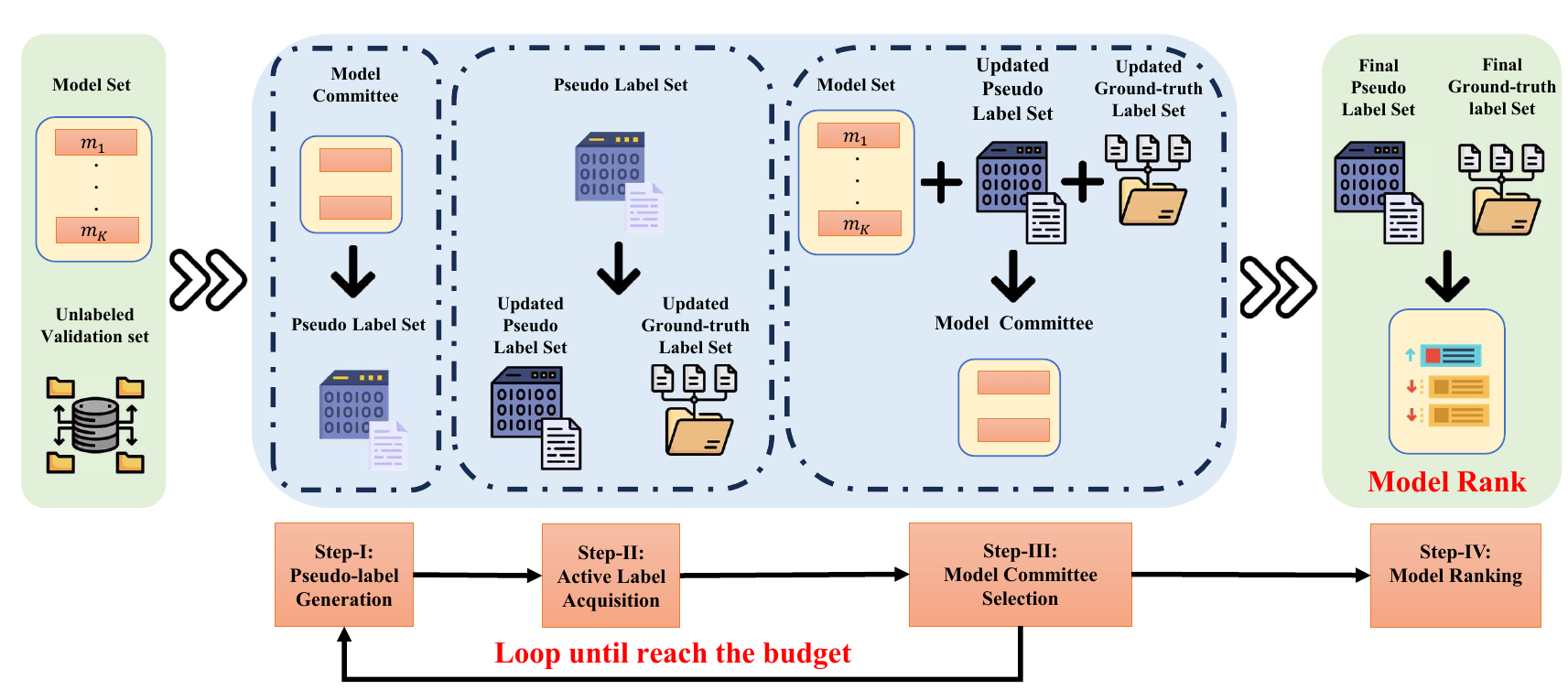}
    \caption{The illustration of the overall procedure of \OURS{}.}
    \label{fig:framework}
\end{figure*}

\section{Methodology}

To rank the trained models, we propose a novel framework \OURS{}, which comprises four primary steps.
\textbf{Step-I}  (Pseudo-label generation, Section~\ref{sec:plg}): Generate pseudo-labels for the unlabeled validation set based on a model committee selected from the model set $\mathcal{M}$.
\textbf{Step-II}  (Active label acquisition, Section~\ref{sec:tlr}): Select samples from the validation set and acquires their ground-truth labels to replace the pseudo-labels.
\textbf{Step-III}  (Model committee selection, Section~\ref{sec:msu}): Select a subset of models based on the updated pseudo-label to form a model committee that would be used to generate pseudo-labels in the next iteration.
After $T$ rounds of iteration for these three steps, we obtain our final pseudo labels, based on which we perform our \textbf{Step-IV}  (Model Ranking, Section~\ref{sec:mr}).
These four steps are detailed in Figure~\ref{fig:framework} and the pseudocode of \OURS{} is shown in Appendix~\ref{alg:alg1}.

\subsection{Step-I: Pseudo-label Generation}
\label{sec:plg} 

Our first step is to generate pseudo-labels based on a subset of trained models referred to as the model committee, which will be introduced soon.
As the trained models usually have a certain level of capability on the task, it is natural to leverage their ensemble to obtain reasonable pseudo-labels~\cite{krogh1994neural,hansen1990neural}.
In particular, we denote $\mathcal{M}_{C}^{t}$ as the model committee at $t$-th iteration, and explore two design choices of pseudo-label generation:

\begin{itemize}

  \item \textbf{Hard ensemble}: For $x_{i} \in D_{V}$, hard ensemble uses the average of the one-hot label prediction vectors generated by all models in $\mathcal{M}_{C}^{t}$ as its pseudo-label distribution $\hat{y}^{(t)}_{i}$.

  \item \textbf{Soft ensemble}: For $x_{i} \in D_{V}$, soft ensemble employs the average of the label probability simplex generated by all models in $\mathcal{M}_{C}^{t}$ as its pseudo-label distribution $\hat{y}^{(t)}_{i}$.

\end{itemize}

Therefore, at $t$-th iteration, we generate the pseudo-label for the $i$-th sample via:
\begin{equation}
\begin{aligned}
\hat{y}^{ (t)}_{i} \leftarrow  g (x_{i}, \mathcal{M}_{C}^{t}).
\end{aligned}
\end{equation}
where the function $g (\cdot)$ could be either hard or soft ensemble.
These pseudo-labels will be used to select high-quality models to form the model committee.

\subsection{Step-II: Active Label Acquisition}
\label{sec:tlr}
In the second step of the \OURS{} framework, we actively acquire labels for a subset of samples from the pseudo-label set.
We explore several existing active sampling strategies in the literature:

\begin{itemize}

  \item \textbf{Random}: Although the random sampling is not part of the uncertainty sampling strategies, as a classical acquisition strategy~\cite{bergstra2012random,rawat2021disentangling}, we also put it into our framework for reference.

  \item \textbf{Uncertainty}~\cite{culotta2005reducing}:  We define the value of 1 minus probabilities of the top predicted class as the uncertainty value for a pseudo-label. 

  \item \textbf{Margin}~\cite{schroder-etal-2022-revisiting}: Here, we target pseudo-labels with the smallest margin between the probabilities of the top two predicted classes.

  \item \textbf{Entropy}~\cite{4563068}: This strategy calculates the entropy for each pseudo-label. With higher entropy indicating higher information, we prioritize acquiring labels with the highest entropy values.

\end{itemize}
Utilizing these strategies, we produce a set $S^{(t)}$ of $b$ samples at each iteration $t$:

\begin{equation}
S^{ (t)}  \leftarrow  l (L_{p}, b) ,
\end{equation}
where $l (\cdot)$ represents a certain acquiring strategy  (Uncertainty, Margin,  Entropy, or Random) and $L_{p}$ is the current set of pseudo-labels. We then acquire ground-truth labels for the selected set $S^{(t)}$.

We denote the set consisting of all ground-truth labels we have acquired as $L_g$. 
For each sample in $S^{(t)}$, we add its ground-truth label to $L_g$ and remove the corresponding pseudo-label from $L_p$. This enhances the reliability of our pseudo-labels and refines subsequent steps, such as model committee selections.

\subsection{Step-III: Model Committee Selection}
\label{sec:msu}

The process of Model Committee Selection in our \OURS{} framework is a critical step to ensure the appropriate models are chosen to produce pseudo-labels for the next iteration. 
In our framework, we explore two distinct methods for model committee selection: Z-score and All-model:

\begin{itemize}

  \item \textbf{All-model}: All-model approach involves utilizing every model in the existing set $\mathcal{M}$ as part of the model committee. 
  It operates on the principle that the ensemble of diverse models can lead to a more generalized and comprehensive understanding, contributing to the robustness of the pseudo-labels generated.

  \item \textbf{Z-score}: The Z-score method assesses a model's performance relative to the median performance of the entire model set $\mathcal{M}$, aiding in the identification and filtering of \emph{outlier models} with extremely low performance. 
  It starts by calculating the accuracy $a_{k}$ of the $k$-th model against the latest pseudo label set $L_p$ and ground-truth label set $L_g$. 
Then, we calculate the Z-score for each model.
Specifically, the Z-score $z_{k}$ of the model $m_k$ is determined as follows:
\begin{equation}
\resizebox{0.75\linewidth}{!}{
$ z_{k} \leftarrow    \frac{\delta \times (a_{k} - a_{m})}{\text{Median} (\{| a_{k^{'}} - a_{m} | : k^{'} \in [K]\})},  $
}
\end{equation}
where $a_{m}$ is the median of the $\{ a_{k} \}_{k \in [K] }$.
Subsequently, models with Z-score exceeding a certain threshold, $\tau$, are selected for the next iteration's committee.
This ensures that only the most predictive and reliable models contribute to the pseudo-label generation.
\end{itemize}

Therefore, at the end of $t$-th iteration, we select the model committee for the $(t+1)$-th iteration as:
\begin{equation}
\mathcal{M}_{C}^{ (t+1)} \leftarrow s (L_p, L_g,\mathcal{M}),
\end{equation}

where the function $s (\cdot)$ could be either Z-score or All-model.
Notably, with the updates of $L_p$ and $L_g$, each time we choose the model committee from all models, not from the last model committee.
This prevents the early exclusion of potentially valuable models, ensuring a robust and dynamic selection process throughout the iterations.

\begin{table*}[t]
\centering
\scalebox{0.40}{

\begin{tabular}{@{}lllccclccclccclccclccclc@{}}
\toprule
\bottomrule
\multicolumn{3}{c|}{\textbf{Method}}                                                                                                                                                                                                                                                                                                                       & \multicolumn{20}{c}{\textbf{Dataset}}                                                                                                                                                                                                                             & \multicolumn{1}{l}{} \\ \midrule
\multicolumn{1}{c|}{\multirow{2}{*}{\textbf{\begin{tabular}[c]{@{}c@{}}Pseudo-label\\ Generation\end{tabular}}}} & \multicolumn{1}{c|}{\multirow{2}{*}{\textbf{\begin{tabular}[c]{@{}c@{}}Active Label\\ Acquisition\end{tabular}}}} & \multicolumn{1}{c|}{\multirow{2}{*}{\textbf{\begin{tabular}[c]{@{}c@{}}Model Committee\\ Selection\end{tabular}}}} & \multicolumn{4}{c|}{\textbf{IMDB (20)}}                     & \multicolumn{4}{c|}{\textbf{AGNews (40)}}                  & \multicolumn{4}{c|}{\textbf{Amazon Review (250)}}           & \multicolumn{4}{c|}{\textbf{Yelp Review (250)}}             & \multicolumn{4}{c|}{\textbf{Yahoo! Answer (500)}}           & \textbf{Avg.}                 \\ \cmidrule(l){4-24} 
\multicolumn{1}{c|}{}                                                                                            & \multicolumn{1}{c|}{}                                                                                             & \multicolumn{1}{c|}{}                                                                                              & 0\%   & 10\%  & 20\%  & \multicolumn{1}{c|}{50\%} & 0\%   & 10\%  & 20\%  & \multicolumn{1}{c|}{50\%} & 0\%   & 10\%  & 20\%  & \multicolumn{1}{c|}{50\%} & 0\%   & 10\%  & 20\%  & \multicolumn{1}{c|}{50\%} & 0\%   & 10\%  & 20\%  & \multicolumn{1}{c|}{50\%}  &                      \\ \midrule
\multicolumn{1}{l|}{\multirow{10}{*}{Hard Ensemble}}                                                              & \multicolumn{1}{l|}{\multirow{2}{*}{Random}}                                                                      & \multicolumn{1}{l|}{All-model}                                                                                     & \multicolumn{1}{c}{0.98} & \multicolumn{1}{c}{0.97} & \multicolumn{1}{c}{1.09} & \multicolumn{1}{c|}{0.76} & \multicolumn{1}{c}{5.38} & \multicolumn{1}{c}{5.35} & \multicolumn{1}{c}{5.31} & \multicolumn{1}{c|}{0.69} & \multicolumn{1}{c}{9.47} & \multicolumn{1}{c}{9.50} & \multicolumn{1}{c}{9.48} & \multicolumn{1}{c|}{9.47} & \multicolumn{1}{c}{14.27} & \multicolumn{1}{c}{14.27} & \multicolumn{1}{c}{14.27} & \multicolumn{1}{c|}{0.62} & \multicolumn{1}{c}{7.11} & \multicolumn{1}{c}{7.01} & \multicolumn{1}{c}{6.82} & \multicolumn{1}{c}{0.93} & \multicolumn{1}{|c}{6.19} \\ \cmidrule(lr){3-3}
\multicolumn{1}{l|}{}                                                                                            & \multicolumn{1}{l|}{}                                                                                             & \multicolumn{1}{l|}{Z-score}                                                                                       & \multicolumn{1}{c}{0.98} & \multicolumn{1}{c}{0.97} & \multicolumn{1}{c}{1.09} & \multicolumn{1}{c|}{0.76} & \multicolumn{1}{c}{5.38} & \multicolumn{1}{c}{5.35} & \multicolumn{1}{c}{5.31} & \multicolumn{1}{c|}{0.69} & \multicolumn{1}{c}{9.47} & \multicolumn{1}{c}{9.50} & \multicolumn{1}{c}{9.48} & \multicolumn{1}{c|}{9.47} & \multicolumn{1}{c}{14.27} & \multicolumn{1}{c}{14.27} & \multicolumn{1}{c}{14.27} & \multicolumn{1}{c|}{0.62} & \multicolumn{1}{c}{7.11} & \multicolumn{1}{c}{7.01} & \multicolumn{1}{c}{6.82} & \multicolumn{1}{c}{0.93} & \multicolumn{1}{|c}{6.19} \\ \cmidrule(lr){2-3}
\multicolumn{1}{l|}{}                                                                                            & \multicolumn{1}{l|}{\multirow{2}{*}{Uncertainty}}                                                                 & \multicolumn{1}{l|}{All-model}                                                                                     & \multicolumn{1}{c}{0.98} & \multicolumn{1}{c}{0.84} & \multicolumn{1}{c}{0.77} & \multicolumn{1}{c|}{0.12} & \multicolumn{1}{c}{5.38} & \multicolumn{1}{c}{4.81} & \multicolumn{1}{c}{0.21} & \multicolumn{1}{c|}{0.01} & \multicolumn{1}{c}{9.47} & \multicolumn{1}{c}{9.48} & \multicolumn{1}{c}{9.54} & \multicolumn{1}{c|}{6.90} & \multicolumn{1}{c}{14.27} & \multicolumn{1}{c}{14.27} & \multicolumn{1}{c}{14.28} & \multicolumn{1}{c|}{0.44} & \multicolumn{1}{c}{7.11} & \multicolumn{1}{c}{6.37} & \multicolumn{1}{c}{1.06} & \multicolumn{1}{c}{0.02} & \multicolumn{1}{|c}{5.32} \\ \cmidrule(lr){3-3}
\multicolumn{1}{l|}{}                                                                                            & \multicolumn{1}{l|}{}                                                                                             & \multicolumn{1}{l|}{Z-score}                                                                                       & \multicolumn{1}{c}{0.98} & \multicolumn{1}{c}{0.24} & \multicolumn{1}{c}{0.00} & \multicolumn{1}{c|}{0.00} & \multicolumn{1}{c}{5.38} & \multicolumn{1}{c}{4.41} & \multicolumn{1}{c}{0.24} & \multicolumn{1}{c|}{0.01} & \multicolumn{1}{c}{9.47} & \multicolumn{1}{c}{9.45} & \multicolumn{1}{c}{9.38} & \multicolumn{1}{c|}{5.66} & \multicolumn{1}{c}{14.27} & \multicolumn{1}{c}{14.30} & \multicolumn{1}{c}{14.28} & \multicolumn{1}{c|}{0.60} & \multicolumn{1}{c}{7.11} & \multicolumn{1}{c}{6.36} & \multicolumn{1}{c}{0.99} & \multicolumn{1}{c}{0.04} & \multicolumn{1}{|c}{5.16} \\ \cmidrule(lr){2-3}
\multicolumn{1}{l|}{}                                                                                            & \multicolumn{1}{l|}{\multirow{2}{*}{Margin}}                                                                      & \multicolumn{1}{l|}{All-model}                                                                                     & \multicolumn{1}{c}{0.98} & \multicolumn{1}{c}{0.84} & \multicolumn{1}{c}{0.77} & \multicolumn{1}{c|}{0.12} & \multicolumn{1}{c}{5.38} & \multicolumn{1}{c}{4.79} & \multicolumn{1}{c}{0.23} & \multicolumn{1}{c|}{0.01} & \multicolumn{1}{c}{9.47} & \multicolumn{1}{c}{9.50} & \multicolumn{1}{c}{9.56} & \multicolumn{1}{c|}{7.34} & \multicolumn{1}{c}{14.27} & \multicolumn{1}{c}{14.27} & \multicolumn{1}{c}{14.28} & \multicolumn{1}{c|}{0.45} & \multicolumn{1}{c}{7.11} & \multicolumn{1}{c}{6.69} & \multicolumn{1}{c}{1.13} & \multicolumn{1}{c}{0.03} & \multicolumn{1}{|c}{5.36} \\
 \cmidrule(lr){3-3}
\multicolumn{1}{l|}{}                                                                                            & \multicolumn{1}{l|}{}                                                                                             & \multicolumn{1}{l|}{Z-score}                                                                                       & \multicolumn{1}{c}{0.98} & \multicolumn{1}{c}{0.24} & \multicolumn{1}{c}{0.00} & \multicolumn{1}{c|}{0.00} & \multicolumn{1}{c}{5.38} & \multicolumn{1}{c}{4.57} & \multicolumn{1}{c}{0.22} & \multicolumn{1}{c|}{0.01} & \multicolumn{1}{c}{9.47} & \multicolumn{1}{c}{9.45} & \multicolumn{1}{c}{9.38} & \multicolumn{1}{c|}{6.52} & \multicolumn{1}{c}{14.27} & \multicolumn{1}{c}{14.31} & \multicolumn{1}{c}{14.26} & \multicolumn{1}{c|}{0.59} & \multicolumn{1}{c}{7.11} & \multicolumn{1}{c}{6.56} & \multicolumn{1}{c}{1.24} & \multicolumn{1}{c}{0.02} & \multicolumn{1}{|c}{5.23} \\ \cmidrule(lr){2-3}
\multicolumn{1}{l|}{}                                                                                            & \multicolumn{1}{l|}{\multirow{2}{*}{Entropy}}                                                                     & \multicolumn{1}{l|}{All-model}                                                                                     & \multicolumn{1}{c}{0.98} & \multicolumn{1}{c}{0.84} & \multicolumn{1}{c}{0.77} & \multicolumn{1}{c|}{0.12} & \multicolumn{1}{c}{5.38} & \multicolumn{1}{c}{4.72} & \multicolumn{1}{c}{0.20} & \multicolumn{1}{c|}{0.01} & \multicolumn{1}{c}{9.47} & \multicolumn{1}{c}{9.50} & \multicolumn{1}{c}{9.56} & \multicolumn{1}{c|}{4.03} & \multicolumn{1}{c}{14.27} & \multicolumn{1}{c}{14.27} & \multicolumn{1}{c}{14.29} & \multicolumn{1}{c|}{0.45} & \multicolumn{1}{c}{7.11} & \multicolumn{1}{c}{6.44} & \multicolumn{1}{c}{0.87} & \multicolumn{1}{c}{0.02} & \multicolumn{1}{|c}{5.17} \\
 \cmidrule(lr){3-3}
\multicolumn{1}{l|}{}                                                                                            & \multicolumn{1}{l|}{}                                                                                             & \multicolumn{1}{l|}{Z-score}                                                                                       & \multicolumn{1}{c}{0.98} & \multicolumn{1}{c}{0.24} & \multicolumn{1}{c}{0.00} & \multicolumn{1}{c|}{0.00} & \multicolumn{1}{c}{5.38} & \multicolumn{1}{c}{4.59} & \multicolumn{1}{c}{0.19} & \multicolumn{1}{c|}{0.01} & \multicolumn{1}{c}{9.47} & \multicolumn{1}{c}{9.45} & \multicolumn{1}{c}{9.43} & \multicolumn{1}{c|}{3.65} & \multicolumn{1}{c}{14.27} & \multicolumn{1}{c}{14.31} & \multicolumn{1}{c}{14.20} & \multicolumn{1}{c|}{0.57} & \multicolumn{1}{c}{7.11} & \multicolumn{1}{c}{6.04} & \multicolumn{1}{c}{0.81} & \multicolumn{1}{c}{0.02} & \multicolumn{1}{|c}{5.04} \\
\toprule
\multicolumn{1}{l|}{\multirow{10}{*}{Soft Ensemble}}                                                              & \multicolumn{1}{l|}{\multirow{2}{*}{Random}}                                                                      & \multicolumn{1}{l|}{All-model}                                                                                     & \multicolumn{1}{c}{1.13} & \multicolumn{1}{c}{1.18} & \multicolumn{1}{c}{1.03} & \multicolumn{1}{c|}{0.76} & \multicolumn{1}{c}{5.41} & \multicolumn{1}{c}{5.35} & \multicolumn{1}{c}{5.31} & \multicolumn{1}{c|}{0.57} & \multicolumn{1}{c}{9.45} & \multicolumn{1}{c}{9.46} & \multicolumn{1}{c}{9.46} & \multicolumn{1}{c|}{9.46} & \multicolumn{1}{c}{14.26} & \multicolumn{1}{c}{14.27} & \multicolumn{1}{c}{14.27} & \multicolumn{1}{c|}{1.51} & \multicolumn{1}{c}{7.11} & \multicolumn{1}{c}{7.01} & \multicolumn{1}{c}{6.77} & \multicolumn{1}{c}{0.93} & \multicolumn{1}{|c}{6.24} \\ \cmidrule(lr){3-3}
\multicolumn{1}{l|}{}                                                                                            & \multicolumn{1}{l|}{}                                                                                             & \multicolumn{1}{l|}{Z-score}                                                                                       & \multicolumn{1}{c}{1.13} & \multicolumn{1}{c}{1.18} & \multicolumn{1}{c}{1.03} & \multicolumn{1}{c|}{0.76} & \multicolumn{1}{c}{5.41} & \multicolumn{1}{c}{5.35} & \multicolumn{1}{c}{5.31} & \multicolumn{1}{c|}{0.57} & \multicolumn{1}{c}{9.45} & \multicolumn{1}{c}{9.46} & \multicolumn{1}{c}{9.46} & \multicolumn{1}{c|}{9.46} & \multicolumn{1}{c}{14.26} & \multicolumn{1}{c}{14.27} & \multicolumn{1}{c}{14.27} & \multicolumn{1}{c|}{1.51} & \multicolumn{1}{c}{7.11} & \multicolumn{1}{c}{7.01} & \multicolumn{1}{c}{6.77} & \multicolumn{1}{c}{0.93} & \multicolumn{1}{|c}{6.24} \\ \cmidrule(lr){2-3}
\multicolumn{1}{l|}{}                                                                                            & \multicolumn{1}{l|}{\multirow{2}{*}{Uncertainty}}                                                                 & \multicolumn{1}{l|}{All-model}                                                                                     & \multicolumn{1}{c}{1.13} & \multicolumn{1}{c}{0.82} & \multicolumn{1}{c}{0.63} & \multicolumn{1}{c|}{0.12} & \multicolumn{1}{c}{5.41} & \multicolumn{1}{c}{4.70} & \multicolumn{1}{c}{0.22} & \multicolumn{1}{c|}{0.02} & \multicolumn{1}{c}{9.45} & \multicolumn{1}{c}{9.47} & \multicolumn{1}{c}{9.48} & \multicolumn{1}{c|}{7.78} & \multicolumn{1}{c}{14.26} & \multicolumn{1}{c}{14.27} & \multicolumn{1}{c}{14.28} & \multicolumn{1}{c|}{0.48} & \multicolumn{1}{c}{7.11} & \multicolumn{1}{c}{6.42} & \multicolumn{1}{c}{1.17} & \multicolumn{1}{c}{0.03} & \multicolumn{1}{|c}{5.36} \\ \cmidrule(lr){3-3}
\multicolumn{1}{l|}{}                                                                                            & \multicolumn{1}{l|}{}                                                                                             & \multicolumn{1}{l|}{Z-score}                                                                                       & \multicolumn{1}{c}{1.13} & \multicolumn{1}{c}{0.34} & \multicolumn{1}{c}{0.02} & \multicolumn{1}{c|}{0.00} & \multicolumn{1}{c}{5.41} & \multicolumn{1}{c}{4.51} & \multicolumn{1}{c}{0.23} & \multicolumn{1}{c|}{0.01} & \multicolumn{1}{c}{9.45} & \multicolumn{1}{c}{9.45} & \multicolumn{1}{c}{9.40} & \multicolumn{1}{c|}{7.91} & \multicolumn{1}{c}{14.26} & \multicolumn{1}{c}{14.27} & \multicolumn{1}{c}{14.27} & \multicolumn{1}{c|}{0.59} & \multicolumn{1}{c}{7.11} & \multicolumn{1}{c}{6.45} & \multicolumn{1}{c}{1.24} & \multicolumn{1}{c}{0.02} & \multicolumn{1}{|c}{5.30} \\ \cmidrule(lr){2-3}
\multicolumn{1}{l|}{}                                                                                            & \multicolumn{1}{l|}{\multirow{2}{*}{Margin}}                                                                      & \multicolumn{1}{l|}{All-model}                                                                                     & \multicolumn{1}{c}{1.13} & \multicolumn{1}{c}{0.82} & \multicolumn{1}{c}{0.63} & \multicolumn{1}{c|}{0.12} & \multicolumn{1}{c}{5.41} & \multicolumn{1}{c}{4.82} & \multicolumn{1}{c}{0.25} & \multicolumn{1}{c|}{0.03} & \multicolumn{1}{c}{9.45} & \multicolumn{1}{c}{9.47} & \multicolumn{1}{c}{9.49} & \multicolumn{1}{c|}{8.09} & \multicolumn{1}{c}{14.26} & \multicolumn{1}{c}{14.27} & \multicolumn{1}{c}{14.28} & \multicolumn{1}{c|}{0.44} & \multicolumn{1}{c}{7.11} & \multicolumn{1}{c}{6.50} & \multicolumn{1}{c}{1.15} & \multicolumn{1}{c}{0.03} & \multicolumn{1}{|c}{5.39} \\ \cmidrule(lr){3-3}
\multicolumn{1}{l|}{}                                                                                            & \multicolumn{1}{l|}{}                                                                                             & \multicolumn{1}{l|}{Z-score}                                                                                       & \multicolumn{1}{c}{1.13} & \multicolumn{1}{c}{0.34} & \multicolumn{1}{c}{0.02} & \multicolumn{1}{c|}{0.00} & \multicolumn{1}{c}{5.41} & \multicolumn{1}{c}{4.29} & \multicolumn{1}{c}{0.21} & \multicolumn{1}{c|}{0.00} & \multicolumn{1}{c}{9.45} & \multicolumn{1}{c}{9.45} & \multicolumn{1}{c}{9.45} & \multicolumn{1}{c|}{8.09} & \multicolumn{1}{c}{14.26} & \multicolumn{1}{c}{14.30} & \multicolumn{1}{c}{14.24} & \multicolumn{1}{c|}{0.64} & \multicolumn{1}{c}{7.11} & \multicolumn{1}{c}{6.55} & \multicolumn{1}{c}{1.12} & \multicolumn{1}{c}{0.04} & \multicolumn{1}{|c}{5.31} \\ \cmidrule(lr){2-3}
\multicolumn{1}{l|}{}                                                                                            & \multicolumn{1}{l|}{\multirow{2}{*}{Entropy}}                                                                     & \multicolumn{1}{l|}{All-model}                                                                                     & \multicolumn{1}{c}{1.13} & \multicolumn{1}{c}{0.82} & \multicolumn{1}{c}{0.63} & \multicolumn{1}{c|}{0.12} & \multicolumn{1}{c}{5.41} & \multicolumn{1}{c}{4.61} & \multicolumn{1}{c}{0.20} & \multicolumn{1}{c|}{0.03} & \multicolumn{1}{c}{9.45} & \multicolumn{1}{c}{9.45} & \multicolumn{1}{c}{9.49} & \multicolumn{1}{c|}{7.10} & \multicolumn{1}{c}{14.26} & \multicolumn{1}{c}{14.27} & \multicolumn{1}{c}{14.27} & \multicolumn{1}{c|}{0.51} & \multicolumn{1}{c}{7.11} & \multicolumn{1}{c}{6.31} & \multicolumn{1}{c}{0.97} & \multicolumn{1}{c}{0.00} & \multicolumn{1}{|c}{5.31} \\ \cmidrule(lr){3-3}
\multicolumn{1}{l|}{}                                                                                            & \multicolumn{1}{l|}{}                                                                                             & \multicolumn{1}{l|}{Z-score}                                                                                                            & \multicolumn{1}{c}{1.13} & \multicolumn{1}{c}{0.34} & \multicolumn{1}{c}{0.02} & \multicolumn{1}{c|}{0.00} & \multicolumn{1}{c}{5.41} & \multicolumn{1}{c}{4.59} & \multicolumn{1}{c}{0.17} & \multicolumn{1}{c|}{0.01} & \multicolumn{1}{c}{9.45} & \multicolumn{1}{c}{9.45} & \multicolumn{1}{c}{9.45} & \multicolumn{1}{c|}{7.15} & \multicolumn{1}{c}{14.26} & \multicolumn{1}{c}{14.31} & \multicolumn{1}{c}{14.28} & \multicolumn{1}{c|}{0.64} & \multicolumn{1}{c}{7.11} & \multicolumn{1}{c}{6.30} & \multicolumn{1}{c}{0.94} & \multicolumn{1}{c}{0.03} & \multicolumn{1}{|c}{5.25} \\

\bottomrule
\bottomrule
\end{tabular}

}
\caption{
\textbf{Semi-supervised learning setting}: 
This table illustrates the changes in optimal gap values within our design space.
These changes are observed across different budget ratios, specifically at 0\%, 10\%, 20\%, and 50\%.
The number in brackets after the dataset indicates the number of labels used in model training stage.
}
\label{tab:usb_og_diff_ratio_t1}
\end{table*}

\begin{table*}[t]
\centering
\scalebox{0.40}{

\begin{tabular}{@{}lllccclccclccclccclccclc@{}}
\toprule
\bottomrule
\multicolumn{3}{c|}{\textbf{Method}}                                                                                                                                                                                                                                                                                                                       & \multicolumn{20}{c}{\textbf{Dataset}}                                                                                                                                                                                                                             & \multicolumn{1}{l}{} \\ \midrule
\multicolumn{1}{c|}{\multirow{2}{*}{\textbf{\begin{tabular}[c]{@{}c@{}}Pseudo-label\\ Generation\end{tabular}}}} & \multicolumn{1}{c|}{\multirow{2}{*}{\textbf{\begin{tabular}[c]{@{}c@{}}Active Label\\ Acquisition\end{tabular}}}} & \multicolumn{1}{c|}{\multirow{2}{*}{\textbf{\begin{tabular}[c]{@{}c@{}}Model Committee\\ Selection\end{tabular}}}} & \multicolumn{4}{c|}{\textbf{IMDB (100)}}                    & \multicolumn{4}{c|}{\textbf{AGNews (200)}}                 & \multicolumn{4}{c|}{\textbf{Amazon Review (1000)}}          & \multicolumn{4}{c|}{\textbf{Yelp Review (1000)}}            & \multicolumn{4}{c|}{\textbf{Yahoo! Answer (2000)}}          & \textbf{Avg.}                 \\ \cmidrule(l){4-24} 
\multicolumn{1}{c|}{}                                                                                            & \multicolumn{1}{c|}{}                                                                                             & \multicolumn{1}{c|}{}                                                                                              & 0\%   & 10\%  & 20\%  & \multicolumn{1}{c|}{50\%} & 0\%   & 10\%  & 20\%  & \multicolumn{1}{c|}{50\%} & 0\%   & 10\%  & 20\%  & \multicolumn{1}{c|}{50\%} & 0\%   & 10\%  & 20\%  & \multicolumn{1}{c|}{50\%} & 0\%   & 10\%  & 20\%  & \multicolumn{1}{c|}{50\%}  &                      \\ \midrule
\multicolumn{1}{l|}{\multirow{10}{*}{Hard Ensemble}}                                                              & \multicolumn{1}{l|}{\multirow{2}{*}{Random}}                                                                      & \multicolumn{1}{l|}{All-model}                                                                                     & \multicolumn{1}{c}{0.96} & \multicolumn{1}{c}{0.96} & \multicolumn{1}{c}{0.91} & \multicolumn{1}{c|}{0.86} & \multicolumn{1}{c}{4.85} & \multicolumn{1}{c}{4.90} & \multicolumn{1}{c}{4.85} & \multicolumn{1}{c|}{2.17} & \multicolumn{1}{c}{7.25} & \multicolumn{1}{c}{7.24} & \multicolumn{1}{c}{7.23} & \multicolumn{1}{c|}{7.20} & \multicolumn{1}{c}{8.64} & \multicolumn{1}{c}{8.65} & \multicolumn{1}{c}{8.65} & \multicolumn{1}{c|}{7.33} & \multicolumn{1}{c}{5.83} & \multicolumn{1}{c}{5.77} & \multicolumn{1}{c}{5.71} & \multicolumn{1}{c}{0.70} & \multicolumn{1}{|c}{5.03} \\  \cmidrule(lr){3-3}
\multicolumn{1}{l|}{}                                                                                            & \multicolumn{1}{l|}{}                                                                                             & \multicolumn{1}{l|}{Z-score}                                                                                       & \multicolumn{1}{c}{0.96} & \multicolumn{1}{c}{0.96} & \multicolumn{1}{c}{0.91} & \multicolumn{1}{c|}{0.86} & \multicolumn{1}{c}{4.85} & \multicolumn{1}{c}{4.90} & \multicolumn{1}{c}{4.85} & \multicolumn{1}{c|}{2.17} & \multicolumn{1}{c}{7.25} & \multicolumn{1}{c}{7.24} & \multicolumn{1}{c}{7.23} & \multicolumn{1}{c|}{7.20} & \multicolumn{1}{c}{8.64} & \multicolumn{1}{c}{8.65} & \multicolumn{1}{c}{8.65} & \multicolumn{1}{c|}{7.33} & \multicolumn{1}{c}{5.83} & \multicolumn{1}{c}{5.77} & \multicolumn{1}{c}{5.71} & \multicolumn{1}{c}{0.70} & \multicolumn{1}{|c}{5.03} \\  \cmidrule(lr){2-3}
\multicolumn{1}{l|}{}                                                                                            & \multicolumn{1}{l|}{\multirow{2}{*}{Uncertainty}}                                                                 & \multicolumn{1}{l|}{All-model}                                                                                     & \multicolumn{1}{c}{0.96} & \multicolumn{1}{c}{0.66} & \multicolumn{1}{c}{0.65} & \multicolumn{1}{c|}{0.06} & \multicolumn{1}{c}{4.85} & \multicolumn{1}{c}{4.28} & \multicolumn{1}{c}{0.05} & \multicolumn{1}{c|}{0.01} & \multicolumn{1}{c}{7.25} & \multicolumn{1}{c}{7.17} & \multicolumn{1}{c}{7.14} & \multicolumn{1}{c|}{2.60} & \multicolumn{1}{c}{8.64} & \multicolumn{1}{c}{8.63} & \multicolumn{1}{c}{8.66} & \multicolumn{1}{c|}{0.36} & \multicolumn{1}{c}{5.83} & \multicolumn{1}{c}{5.39} & \multicolumn{1}{c}{0.70} & \multicolumn{1}{c}{0.01} & \multicolumn{1}{|c}{3.70} \\  \cmidrule(lr){3-3}
\multicolumn{1}{l|}{}                                                                                            & \multicolumn{1}{l|}{}                                                                                             & \multicolumn{1}{l|}{Z-score}                                                                                       & \multicolumn{1}{c}{0.96} & \multicolumn{1}{c}{0.67} & \multicolumn{1}{c}{0.04} & \multicolumn{1}{c|}{0.00} & \multicolumn{1}{c}{4.85} & \multicolumn{1}{c}{4.47} & \multicolumn{1}{c}{0.04} & \multicolumn{1}{c|}{0.00} & \multicolumn{1}{c}{7.25} & \multicolumn{1}{c}{7.24} & \multicolumn{1}{c}{7.03} & \multicolumn{1}{c|}{2.70} & \multicolumn{1}{c}{8.64} & \multicolumn{1}{c}{8.71} & \multicolumn{1}{c}{8.67} & \multicolumn{1}{c|}{0.34} & \multicolumn{1}{c}{5.83} & \multicolumn{1}{c}{5.26} & \multicolumn{1}{c}{0.66} & \multicolumn{1}{c}{0.01} & \multicolumn{1}{|c}{3.67} \\  \cmidrule(lr){2-3}
\multicolumn{1}{l|}{}                                                                                            & \multicolumn{1}{l|}{\multirow{2}{*}{Margin}}                                                                      & \multicolumn{1}{l|}{All-model}                                                                                     & \multicolumn{1}{c}{0.96} & \multicolumn{1}{c}{0.66} & \multicolumn{1}{c}{0.65} & \multicolumn{1}{c|}{0.06} & \multicolumn{1}{c}{4.85} & \multicolumn{1}{c}{4.48} & \multicolumn{1}{c}{0.07} & \multicolumn{1}{c|}{0.01} & \multicolumn{1}{c}{7.25} & \multicolumn{1}{c}{7.19} & \multicolumn{1}{c}{7.10} & \multicolumn{1}{c|}{2.62} & \multicolumn{1}{c}{8.64} & \multicolumn{1}{c}{8.65} & \multicolumn{1}{c}{8.70} & \multicolumn{1}{c|}{0.39} & \multicolumn{1}{c}{5.83} & \multicolumn{1}{c}{5.24} & \multicolumn{1}{c}{0.87} & \multicolumn{1}{c}{0.00} & \multicolumn{1}{|c}{3.71} \\  \cmidrule(lr){3-3}
\multicolumn{1}{l|}{}                                                                                            & \multicolumn{1}{l|}{}                                                                                             & \multicolumn{1}{l|}{Z-score}                                                                                        & \multicolumn{1}{c}{0.96} & \multicolumn{1}{c}{0.67} & \multicolumn{1}{c}{0.04} & \multicolumn{1}{c|}{0.00} & \multicolumn{1}{c}{4.85} & \multicolumn{1}{c}{4.44} & \multicolumn{1}{c}{0.05} & \multicolumn{1}{c|}{0.00} & \multicolumn{1}{c}{7.25} & \multicolumn{1}{c}{7.26} & \multicolumn{1}{c}{7.17} & \multicolumn{1}{c|}{2.83} & \multicolumn{1}{c}{8.64} & \multicolumn{1}{c}{8.69} & \multicolumn{1}{c}{8.69} & \multicolumn{1}{c|}{0.31} & \multicolumn{1}{c}{5.83} & \multicolumn{1}{c}{5.21} & \multicolumn{1}{c}{0.85} & \multicolumn{1}{c}{0.00} & \multicolumn{1}{|c}{3.69} \\  \cmidrule(lr){2-3}
\multicolumn{1}{l|}{}                                                                                            & \multicolumn{1}{l|}{\multirow{2}{*}{Entropy}}                                                                     & \multicolumn{1}{l|}{All-model}                                                                                     & \multicolumn{1}{c}{0.96} & \multicolumn{1}{c}{0.66} & \multicolumn{1}{c}{0.65} & \multicolumn{1}{c|}{0.06} & \multicolumn{1}{c}{4.85} & \multicolumn{1}{c}{3.92} & \multicolumn{1}{c}{0.04} & \multicolumn{1}{c|}{0.01} & \multicolumn{1}{c}{7.25} & \multicolumn{1}{c}{7.18} & \multicolumn{1}{c}{7.04} & \multicolumn{1}{c|}{1.86} & \multicolumn{1}{c}{8.64} & \multicolumn{1}{c}{8.65} & \multicolumn{1}{c}{8.67} & \multicolumn{1}{c|}{0.42} & \multicolumn{1}{c}{5.83} & \multicolumn{1}{c}{5.40} & \multicolumn{1}{c}{0.60} & \multicolumn{1}{c}{0.01} & \multicolumn{1}{|c}{3.63} \\  \cmidrule(lr){3-3}
\multicolumn{1}{l|}{}                                                                                            & \multicolumn{1}{l|}{}                                                                                             & \multicolumn{1}{l|}{Z-score}                                                                                       & \multicolumn{1}{c}{0.96} & \multicolumn{1}{c}{0.67} & \multicolumn{1}{c}{0.04} & \multicolumn{1}{c|}{0.00} & \multicolumn{1}{c}{4.85} & \multicolumn{1}{c}{3.92} & \multicolumn{1}{c}{0.04} & \multicolumn{1}{c|}{0.00} & \multicolumn{1}{c}{7.25} & \multicolumn{1}{c}{7.26} & \multicolumn{1}{c}{7.08} & \multicolumn{1}{c|}{2.37} & \multicolumn{1}{c}{8.64} & \multicolumn{1}{c}{8.70} & \multicolumn{1}{c}{8.64} & \multicolumn{1}{c|}{0.34} & \multicolumn{1}{c}{5.83} & \multicolumn{1}{c}{5.30} & \multicolumn{1}{c}{0.60} & \multicolumn{1}{c}{0.01} & \multicolumn{1}{|c}{3.63} \\  \toprule
\multicolumn{1}{l|}{\multirow{10}{*}{Soft Ensemble}}                                                              & \multicolumn{1}{l|}{\multirow{2}{*}{Random}}                                                                      & \multicolumn{1}{l|}{All-model}                                                                                     & \multicolumn{1}{c}{0.99} & \multicolumn{1}{c}{0.94} & \multicolumn{1}{c}{0.95} & \multicolumn{1}{c|}{0.91} & \multicolumn{1}{c}{4.88} & \multicolumn{1}{c}{4.87} & \multicolumn{1}{c}{4.88} & \multicolumn{1}{c|}{2.29} & \multicolumn{1}{c}{7.25} & \multicolumn{1}{c}{7.25} & \multicolumn{1}{c}{7.25} & \multicolumn{1}{c|}{7.24} & \multicolumn{1}{c}{8.68} & \multicolumn{1}{c}{8.69} & \multicolumn{1}{c}{8.67} & \multicolumn{1}{c|}{8.16} & \multicolumn{1}{c}{5.83} & \multicolumn{1}{c}{5.82} & \multicolumn{1}{c}{5.67} & \multicolumn{1}{c}{0.66} & \multicolumn{1}{|c}{5.09} \\  \cmidrule(lr){3-3}
\multicolumn{1}{l|}{}                                                                                            & \multicolumn{1}{l|}{}                                                                                             & \multicolumn{1}{l|}{Z-score}                                                                                       & \multicolumn{1}{c}{0.99} & \multicolumn{1}{c}{0.94} & \multicolumn{1}{c}{0.95} & \multicolumn{1}{c|}{0.91} & \multicolumn{1}{c}{4.88} & \multicolumn{1}{c}{4.87} & \multicolumn{1}{c}{4.88} & \multicolumn{1}{c|}{2.29} & \multicolumn{1}{c}{7.25} & \multicolumn{1}{c}{7.25} & \multicolumn{1}{c}{7.25} & \multicolumn{1}{c|}{7.24} & \multicolumn{1}{c}{8.68} & \multicolumn{1}{c}{8.69} & \multicolumn{1}{c}{8.67} & \multicolumn{1}{c|}{8.16} & \multicolumn{1}{c}{5.83} & \multicolumn{1}{c}{5.82} & \multicolumn{1}{c}{5.67} & \multicolumn{1}{c}{0.66} & \multicolumn{1}{|c}{5.09} \\  \cmidrule(lr){2-3}
\multicolumn{1}{l|}{}                                                                                            & \multicolumn{1}{l|}{\multirow{2}{*}{Uncertainty}}                                                                 & \multicolumn{1}{l|}{All-model}                                                                                     & \multicolumn{1}{c}{0.99} & \multicolumn{1}{c}{0.68} & \multicolumn{1}{c}{0.60} & \multicolumn{1}{c|}{0.08} & \multicolumn{1}{c}{4.88} & \multicolumn{1}{c}{4.37} & \multicolumn{1}{c}{0.05} & \multicolumn{1}{c|}{0.01} & \multicolumn{1}{c}{7.25} & \multicolumn{1}{c}{7.18} & \multicolumn{1}{c}{7.12} & \multicolumn{1}{c|}{4.74} & \multicolumn{1}{c}{8.68} & \multicolumn{1}{c}{8.66} & \multicolumn{1}{c}{8.67} & \multicolumn{1}{c|}{0.36} & \multicolumn{1}{c}{5.83} & \multicolumn{1}{c}{5.41} & \multicolumn{1}{c}{0.86} & \multicolumn{1}{c}{0.01} & \multicolumn{1}{|c}{3.82} \\  \cmidrule(lr){3-3}
\multicolumn{1}{l|}{}                                                                                            & \multicolumn{1}{l|}{}                                                                                             & \multicolumn{1}{l|}{Z-score}                                                                                       & \multicolumn{1}{c}{0.99} & \multicolumn{1}{c}{0.67} & \multicolumn{1}{c}{0.16} & \multicolumn{1}{c|}{0.00} & \multicolumn{1}{c}{4.88} & \multicolumn{1}{c}{4.39} & \multicolumn{1}{c}{0.05} & \multicolumn{1}{c|}{0.02} & \multicolumn{1}{c}{7.25} & \multicolumn{1}{c}{7.25} & \multicolumn{1}{c}{7.12} & \multicolumn{1}{c|}{4.72} & \multicolumn{1}{c}{8.68} & \multicolumn{1}{c}{8.69} & \multicolumn{1}{c}{8.70} & \multicolumn{1}{c|}{0.31} & \multicolumn{1}{c}{5.83} & \multicolumn{1}{c}{5.42} & \multicolumn{1}{c}{0.86} & \multicolumn{1}{c}{0.01} & \multicolumn{1}{|c}{3.80} \\
 \cmidrule(lr){2-3}
\multicolumn{1}{l|}{}                                                                                            & \multicolumn{1}{l|}{\multirow{2}{*}{Margin}}                                                                      & \multicolumn{1}{l|}{All-model}                                                                                     & \multicolumn{1}{c}{0.99} & \multicolumn{1}{c}{0.68} & \multicolumn{1}{c}{0.60} & \multicolumn{1}{c|}{0.08} & \multicolumn{1}{c}{4.88} & \multicolumn{1}{c}{4.52} & \multicolumn{1}{c}{0.05} & \multicolumn{1}{c|}{0.00} & \multicolumn{1}{c}{7.25} & \multicolumn{1}{c}{7.23} & \multicolumn{1}{c}{7.16} & \multicolumn{1}{c|}{5.28} & \multicolumn{1}{c}{8.68} & \multicolumn{1}{c}{8.67} & \multicolumn{1}{c}{8.69} & \multicolumn{1}{c|}{0.35} & \multicolumn{1}{c}{5.83} & \multicolumn{1}{c}{5.29} & \multicolumn{1}{c}{0.90} & \multicolumn{1}{c}{0.00} & \multicolumn{1}{|c}{3.86} \\  \cmidrule(lr){3-3}
\multicolumn{1}{l|}{}                                                                                            & \multicolumn{1}{l|}{}                                                                                             & \multicolumn{1}{l|}{Z-score}                                                                                       & \multicolumn{1}{c}{0.99} & \multicolumn{1}{c}{0.67} & \multicolumn{1}{c}{0.16} & \multicolumn{1}{c|}{0.00} & \multicolumn{1}{c}{4.88} & \multicolumn{1}{c}{4.47} & \multicolumn{1}{c}{0.04} & \multicolumn{1}{c|}{0.02} & \multicolumn{1}{c}{7.25} & \multicolumn{1}{c}{7.24} & \multicolumn{1}{c}{7.18} & \multicolumn{1}{c|}{5.11} & \multicolumn{1}{c}{8.68} & \multicolumn{1}{c}{8.70} & \multicolumn{1}{c}{8.73} & \multicolumn{1}{c|}{0.29} & \multicolumn{1}{c}{5.83} & \multicolumn{1}{c}{5.29} & \multicolumn{1}{c}{0.98} & \multicolumn{1}{c}{0.01} & \multicolumn{1}{|c}{3.83} \\  \cmidrule(lr){2-3}
\multicolumn{1}{l|}{}                                                                                            & \multicolumn{1}{l|}{\multirow{2}{*}{Entropy}}                                                                     & \multicolumn{1}{l|}{All-model}                                                                                     & \multicolumn{1}{c}{0.99} & \multicolumn{1}{c}{0.68} & \multicolumn{1}{c}{0.60} & \multicolumn{1}{c|}{0.08} & \multicolumn{1}{c}{4.88} & \multicolumn{1}{c}{4.30} & \multicolumn{1}{c}{0.05} & \multicolumn{1}{c|}{0.01} & \multicolumn{1}{c}{7.25} & \multicolumn{1}{c}{7.18} & \multicolumn{1}{c}{7.13} & \multicolumn{1}{c|}{4.85} & \multicolumn{1}{c}{8.68} & \multicolumn{1}{c}{8.66} & \multicolumn{1}{c}{8.64} & \multicolumn{1}{c|}{0.41} & \multicolumn{1}{c}{5.83} & \multicolumn{1}{c}{5.54} & \multicolumn{1}{c}{0.61} & \multicolumn{1}{c}{0.00} & \multicolumn{1}{|c}{3.82} \\
 \cmidrule(lr){3-3}
\multicolumn{1}{l|}{}                                                                                            & \multicolumn{1}{l|}{}                                                                                             & \multicolumn{1}{l|}{Z-score}                                                                                         & \multicolumn{1}{c}{0.99} & \multicolumn{1}{c}{0.67} & \multicolumn{1}{c}{0.16} & \multicolumn{1}{c|}{0.00} & \multicolumn{1}{c}{4.88} & \multicolumn{1}{c}{4.35} & \multicolumn{1}{c}{0.03} & \multicolumn{1}{c|}{0.02} & \multicolumn{1}{c}{7.25} & \multicolumn{1}{c}{7.19} & \multicolumn{1}{c}{7.15} & \multicolumn{1}{c|}{4.55} & \multicolumn{1}{c}{8.68} & \multicolumn{1}{c}{8.72} & \multicolumn{1}{c}{8.70} & \multicolumn{1}{c|}{0.35} & \multicolumn{1}{c}{5.83} & \multicolumn{1}{c}{5.56} & \multicolumn{1}{c}{0.58} & \multicolumn{1}{c}{0.00} & \multicolumn{1}{|c}{3.78} \\ 
\bottomrule
\bottomrule
\end{tabular}

}
\caption{\textbf{Semi-supervised learning setting}: 
This table illustrates the changes in optimal gap values within our design space.
These changes are observed across different budget ratios, specifically at 0\%, 10\%, 20\%, and 50\%.
The number in brackets after the dataset indicates the number of labels used in model training stage.
}
\label{tab:usb_og_diff_ratio_t2}
\end{table*}

\begin{table*}[t]
\centering
\scalebox{0.5}{
\begin{tabular}{@{}llllllllllllll@{}}
\toprule
\bottomrule
\multicolumn{3}{c|}{\textbf{Method}}                                                                                                                                                                                                                                    & \multicolumn{10}{c}{\textbf{Dataset}}                                                                                                                                                                                                                                                          & \multicolumn{1}{c}{}      \\ \midrule
\multicolumn{1}{c}{\multirow{2}{*}{\begin{tabular}[c]{@{}c@{}}\textbf{Pseudo-label}\\ \textbf{Generation}\end{tabular}}} & \multicolumn{1}{c}{\multirow{2}{*}{\begin{tabular}[c]{@{}c@{}}\textbf{Active Label}\\ \textbf{Acquisition}\end{tabular}}} & \multicolumn{1}{c|}{\multirow{2}{*}{\begin{tabular}[c]{@{}c@{}}\textbf{Model Committee}\\ \textbf{Selection}\end{tabular}}} & \multicolumn{2}{c|}{\textbf{IMDB}}                             & \multicolumn{2}{c|}{\textbf{AGNews}}                          & \multicolumn{2}{c|}{\textbf{Amazon Review}}                    & \multicolumn{2}{c|}{\textbf{Yelp Review}}                    & \multicolumn{2}{c|}{\textbf{Yahoo! Answer}}                      & \multicolumn{1}{c}{\textbf{Avg.}}  \\ \cmidrule (l){4-14} 
\multicolumn{1}{c}{}                                                               & \multicolumn{1}{c}{}                                                               & \multicolumn{1}{c|}{}                                                               & \multicolumn{1}{c}{20}    & \multicolumn{1}{c|}{100}  & \multicolumn{1}{c}{40}    & \multicolumn{1}{c|}{200}  & \multicolumn{1}{c}{250}   & \multicolumn{1}{c|}{1000} & \multicolumn{1}{c}{250}   & \multicolumn{1}{c|}{1000} & \multicolumn{1}{c}{500}   & \multicolumn{1}{c|}{2000} & \multicolumn{1}{c}{}      \\ \midrule
\multicolumn{1}{l|}{\multirow{10}{*}{Hard Ensemble}}                                & \multicolumn{1}{l|}{Random}                                                        & \multicolumn{1}{l|}{All-model}                                                       & \multicolumn{1}{c}{396} & \multicolumn{1}{c}{399} & \multicolumn{1}{c}{1442} & \multicolumn{1}{c}{1321} & \multicolumn{1}{c}{4230} & \multicolumn{1}{c}{4511} & \multicolumn{1}{c}{4363} & \multicolumn{1}{c}{3740} & \multicolumn{1}{c}{6865} & \multicolumn{1}{c|}{7806} & \multicolumn{1}{c}{3507.7} \\    \cmidrule (lr){3-3}
\multicolumn{1}{l|}{}                                                              & \multicolumn{1}{l|}{}                                                              & \multicolumn{1}{l|}{Z-score}                                                        & \multicolumn{1}{c}{396} & \multicolumn{1}{c}{399} & \multicolumn{1}{c}{1442} & \multicolumn{1}{c}{1321} & \multicolumn{1}{c}{4230} & \multicolumn{1}{c}{4511} & \multicolumn{1}{c}{4363} & \multicolumn{1}{c}{3740} & \multicolumn{1}{c}{6865} & \multicolumn{1}{c|}{7806} & \multicolumn{1}{c}{3507.7} \\     \cmidrule (lr){2-3}
\multicolumn{1}{l|}{}                                                              & \multicolumn{1}{l|}{\multirow{2}{*}{Uncertainty}}                                  & \multicolumn{1}{l|}{All-model}                                                      & \multicolumn{1}{c}{239} & \multicolumn{1}{c}{277} & \multicolumn{1}{c}{672} & \multicolumn{1}{c}{393} & \multicolumn{1}{c}{3984} & \multicolumn{1}{c}{3495} & \multicolumn{1}{c}{3959} & \multicolumn{1}{c}{3285} & \multicolumn{1}{c}{3304} & \multicolumn{1}{c|}{3829} & \multicolumn{1}{c}{2344.1} \\   \cmidrule (lr){3-3}
\multicolumn{1}{l|}{}                                                              & \multicolumn{1}{l|}{}                                                              & \multicolumn{1}{l|}{Z-score}                                                        & \multicolumn{1}{c}{57} & \multicolumn{1}{c}{97} & \multicolumn{1}{c}{668} & \multicolumn{1}{c}{392} & \multicolumn{1}{c}{3896} & \multicolumn{1}{c}{3355} & \multicolumn{1}{c}{3941} & \multicolumn{1}{c}{3107} & \multicolumn{1}{c}{3301} & \multicolumn{1}{c|}{3829} & \multicolumn{1}{c}{2264.7} \\     \cmidrule (lr){2-3}
\multicolumn{1}{l|}{}                                                              & \multicolumn{1}{l|}{\multirow{2}{*}{Margin}}                                       & \multicolumn{1}{l|}{All-model}                                                      & \multicolumn{1}{c}{239} & \multicolumn{1}{c}{277} & \multicolumn{1}{c}{667} & \multicolumn{1}{c}{396} & \multicolumn{1}{c}{4057} & \multicolumn{1}{c}{3385} & \multicolumn{1}{c}{4137} & \multicolumn{1}{c}{3349} & \multicolumn{1}{c}{3336} & \multicolumn{1}{c|}{3819} & \multicolumn{1}{c}{2366.8} \\   \cmidrule (lr){3-3}
\multicolumn{1}{l|}{}                                                              & \multicolumn{1}{l|}{}                                                              & \multicolumn{1}{l|}{Z-score}                                                       & \multicolumn{1}{c}{57} & \multicolumn{1}{c}{97} & \multicolumn{1}{c}{671} & \multicolumn{1}{c}{391} & \multicolumn{1}{c}{3914} & \multicolumn{1}{c}{3369} & \multicolumn{1}{c}{3954} & \multicolumn{1}{c}{3124} & \multicolumn{1}{c}{3326} & \multicolumn{1}{c|}{3879} & \multicolumn{1}{c}{2278.4} \\     \cmidrule (lr){2-3}
\multicolumn{1}{l|}{}                                                              & \multicolumn{1}{l|}{\multirow{2}{*}{Entropy}}                                      & \multicolumn{1}{l|}{All-model}                                                      & \multicolumn{1}{c}{239} & \multicolumn{1}{c}{277} & \multicolumn{1}{c}{668} & \multicolumn{1}{c}{387} & \multicolumn{1}{c}{3969} & \multicolumn{1}{c}{3586} & \multicolumn{1}{c}{3902} & \multicolumn{1}{c}{3382} & \multicolumn{1}{c}{3194} & \multicolumn{1}{c|}{3813} & \multicolumn{1}{c}{2342.1} \\   \cmidrule (lr){3-3}
\multicolumn{1}{l|}{}                                                              & \multicolumn{1}{l|}{}                                                              & \multicolumn{1}{l|}{Z-score}                                                        & \multicolumn{1}{c}{57} & \multicolumn{1}{c}{97} & \multicolumn{1}{c}{665} & \multicolumn{1}{c}{393} & \multicolumn{1}{c}{3881} & \multicolumn{1}{c}{3318} & \multicolumn{1}{c}{3919} & \multicolumn{1}{c}{3202} & \multicolumn{1}{c}{2959} & \multicolumn{1}{c|}{3906} & \multicolumn{1}{c}{2240.0} \\     \toprule
\multicolumn{1}{l|}{\multirow{10}{*}{Soft Ensemble}}                                & \multicolumn{1}{l|}{\multirow{2}{*}{Random}}                                       & \multicolumn{1}{l|}{All-model}                                                      & \multicolumn{1}{c}{396} & \multicolumn{1}{c}{399} & \multicolumn{1}{c}{1392} & \multicolumn{1}{c}{1291} & \multicolumn{1}{c}{4236} & \multicolumn{1}{c}{4523} & \multicolumn{1}{c}{4394} & \multicolumn{1}{c}{3860} & \multicolumn{1}{c}{7306} & \multicolumn{1}{c|}{7805} & \multicolumn{1}{c}{3560.5} \\   \cmidrule (lr){3-3}
\multicolumn{1}{l|}{}                                                              & \multicolumn{1}{l|}{}                                                              & \multicolumn{1}{l|}{Z-score}                                                        & \multicolumn{1}{c}{396} & \multicolumn{1}{c}{399} & \multicolumn{1}{c}{1392} & \multicolumn{1}{c}{1291} & \multicolumn{1}{c}{4236} & \multicolumn{1}{c}{4523} & \multicolumn{1}{c}{4394} & \multicolumn{1}{c}{3860} & \multicolumn{1}{c}{7306} & \multicolumn{1}{c|}{7805} & \multicolumn{1}{c}{3560.5} \\     \cmidrule (lr){2-3}
\multicolumn{1}{l|}{}                                                              & \multicolumn{1}{l|}{\multirow{2}{*}{Uncertainty}}                                  & \multicolumn{1}{l|}{All-model}                                                      & \multicolumn{1}{c}{219} & \multicolumn{1}{c}{277} & \multicolumn{1}{c}{709} & \multicolumn{1}{c}{395} & \multicolumn{1}{c}{4078} & \multicolumn{1}{c}{3546} & \multicolumn{1}{c}{3964} & \multicolumn{1}{c}{3369} & \multicolumn{1}{c}{3342} & \multicolumn{1}{c|}{3950} & \multicolumn{1}{c}{2385.3} \\   \cmidrule (lr){3-3}
\multicolumn{1}{l|}{}                                                              & \multicolumn{1}{l|}{}                                                              & \multicolumn{1}{l|}{Z-score}                                                        & \multicolumn{1}{c}{89} & \multicolumn{1}{c}{99} & \multicolumn{1}{c}{650} & \multicolumn{1}{c}{395} & \multicolumn{1}{c}{4026} & \multicolumn{1}{c}{3486} & \multicolumn{1}{c}{3961} & \multicolumn{1}{c}{3247} & \multicolumn{1}{c}{3320} & \multicolumn{1}{c|}{3970} & \multicolumn{1}{c}{2324.6} \\     \cmidrule (lr){2-3}
\multicolumn{1}{l|}{}                                                              & \multicolumn{1}{l|}{\multirow{2}{*}{Margin}}                                       & \multicolumn{1}{l|}{All-model}                                                      & \multicolumn{1}{c}{219} & \multicolumn{1}{c}{277} & \multicolumn{1}{c}{692} & \multicolumn{1}{c}{394} & \multicolumn{1}{c}{4110} & \multicolumn{1}{c}{3511} & \multicolumn{1}{c}{4152} & \multicolumn{1}{c}{3364} & \multicolumn{1}{c}{3397} & \multicolumn{1}{c|}{3902} & \multicolumn{1}{c}{2402.1} \\   \cmidrule (lr){3-3}
\multicolumn{1}{l|}{}                                                              & \multicolumn{1}{l|}{}                                                              & \multicolumn{1}{l|}{Z-score}                                                        & \multicolumn{1}{c}{89} & \multicolumn{1}{c}{99} & \multicolumn{1}{c}{669} & \multicolumn{1}{c}{393} & \multicolumn{1}{c}{3968} & \multicolumn{1}{c}{3652} & \multicolumn{1}{c}{3979} & \multicolumn{1}{c}{3249} & \multicolumn{1}{c}{3364} & \multicolumn{1}{c|}{3907} & \multicolumn{1}{c}{2337.2} \\     \cmidrule (lr){2-3}
\multicolumn{1}{l|}{}                                                              & \multicolumn{1}{l|}{\multirow{2}{*}{Entropy}}                                      & \multicolumn{1}{l|}{All-model}                                                      & \multicolumn{1}{c}{219} & \multicolumn{1}{c}{277} & \multicolumn{1}{c}{683} & \multicolumn{1}{c}{395} & \multicolumn{1}{c}{4006} & \multicolumn{1}{c}{3448} & \multicolumn{1}{c}{3952} & \multicolumn{1}{c}{3422} & \multicolumn{1}{c}{3272} & \multicolumn{1}{c|}{3907} & \multicolumn{1}{c}{2358.6} \\    \cmidrule (lr){3-3}
\multicolumn{1}{l|}{}                                                              & \multicolumn{1}{l|}{}                                                              & \multicolumn{1}{l|}{Z-score}                                                      & \multicolumn{1}{c}{89} & \multicolumn{1}{c}{99} & \multicolumn{1}{c}{673} & \multicolumn{1}{c}{394} & \multicolumn{1}{c}{3943} & \multicolumn{1}{c}{3604} & \multicolumn{1}{c}{3924} & \multicolumn{1}{c}{3272} & \multicolumn{1}{c}{3261} & \multicolumn{1}{c|}{3975} & \multicolumn{1}{c}{2323.8} \\

\toprule
\multicolumn{3}{c|}{\textbf{Size of Unlabeled Validation Set} $D_{V}$} & \multicolumn{1}{c}{400} & \multicolumn{1}{c}{400} & \multicolumn{1}{c}{2000} & \multicolumn{1}{c}{2000} & \multicolumn{1}{c}{5000} & \multicolumn{1}{c}{5000} & \multicolumn{1}{c}{5000} & \multicolumn{1}{c}{5000} & \multicolumn{1}{c}{10000} & \multicolumn{1}{c|}{10000} & \multicolumn{1}{c}{-} \\ 

\bottomrule
\bottomrule

\end{tabular}

}
\caption{\textbf{Semi-supervised learning setting}: 
This table illustrates the minimum labeling budget necessary to achieve an optimal gap of zero in our framework.
The number under the dataset indicates the number of labels used in model training stage.
}
\label{tab:usb_optimal_gap}
\end{table*}

\begin{figure*}[t]
\centering
\includegraphics[scale=0.17]{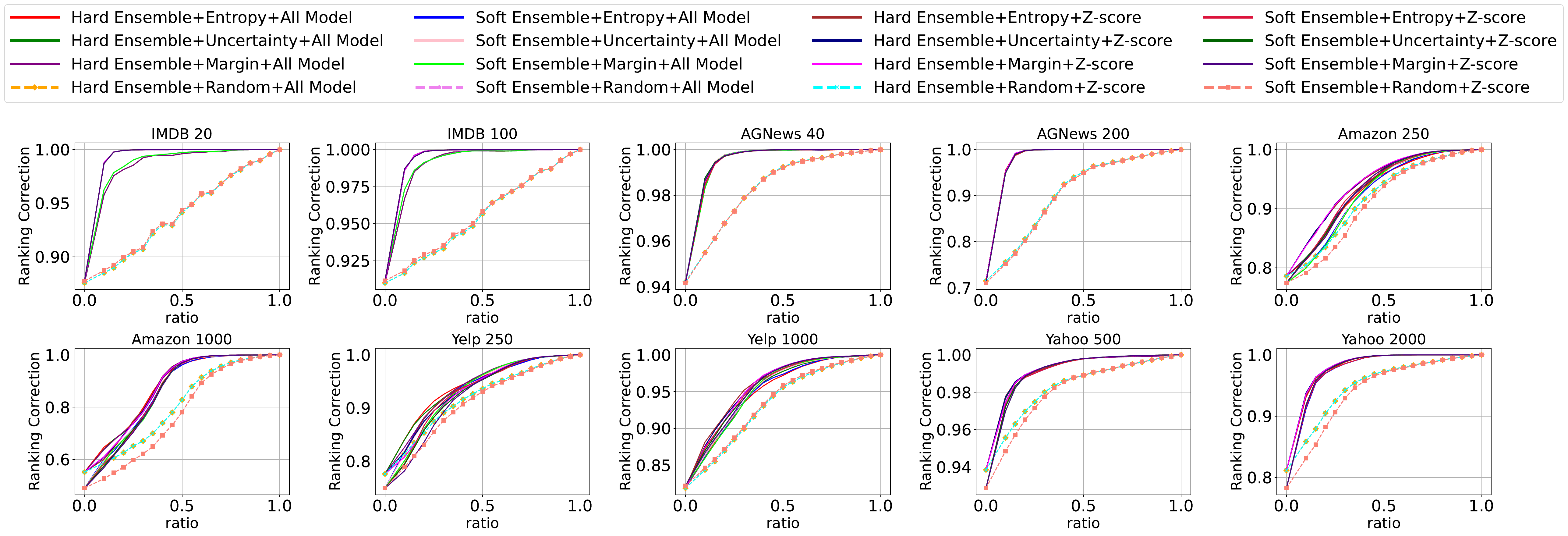}
\caption{\textbf{Semi-supervised learning setting}: 
This figure illustrates the changes in ranking correction values within our design space. 
These changes are observed across budget ratios from 0 to 1.
The number after the dataset indicates the number of labels under the model training stage.
}
\vspace{-1ex}
\label{fig:usb_rc}
\end{figure*}

\begin{table*}[t]
\centering
\scalebox{0.44}{

\begin{tabular}{@{}lllccclccclccclccclccclc@{}}
\toprule
\bottomrule
\multicolumn{3}{c|}{\textbf{Method}}                                                                                                                                                                                                                                                                                                                       & \multicolumn{20}{c}{\textbf{Dataset}}                                                                                                                                                                                                                             & \multicolumn{1}{l}{} \\ \midrule
\multicolumn{1}{c|}{\multirow{2}{*}{\textbf{\begin{tabular}[c]{@{}c@{}}Pseudo-label\\ Generation\end{tabular}}}} & \multicolumn{1}{c|}{\multirow{2}{*}{\textbf{\begin{tabular}[c]{@{}c@{}}Active Label\\ Acquisition\end{tabular}}}} & \multicolumn{1}{c|}{\multirow{2}{*}{\textbf{\begin{tabular}[c]{@{}c@{}}Model Committee\\ Selection\end{tabular}}}} & \multicolumn{4}{c|}{\textbf{Yelp}}                         & \multicolumn{4}{c|}{\textbf{SMS}}                          & \multicolumn{4}{c|}{\textbf{IMDB}}                         & \multicolumn{4}{c|}{\textbf{AGNews}}                       & \multicolumn{4}{c|}{\textbf{Trec}}                         & \textbf{Avg.}                 \\ \cmidrule(l){4-24} 
\multicolumn{1}{c|}{}                                                                                            & \multicolumn{1}{c|}{}                                                                                             & \multicolumn{1}{c|}{}                                                                                              & 0\%   & 10\%  & 20\%  & \multicolumn{1}{c|}{50\%} & 0\%   & 10\%  & 20\%  & \multicolumn{1}{c|}{50\%} & 0\%   & 10\%  & 20\%  & \multicolumn{1}{c|}{50\%} & 0\%   & 10\%  & 20\%  & \multicolumn{1}{c|}{50\%} & 0\%   & 10\%  & 20\%  & \multicolumn{1}{c|}{50\%}  &                      \\ \midrule
\multicolumn{1}{l|}{\multirow{10}{*}{Hard Ensemble}}                                                              & \multicolumn{1}{l|}{\multirow{2}{*}{Random}}                                                                      & \multicolumn{1}{l|}{All-model}                                                                                     & \multicolumn{1}{c}{22.27} & \multicolumn{1}{c}{21.50} & \multicolumn{1}{c}{20.56} & \multicolumn{1}{c|}{13.50} & \multicolumn{1}{c}{0.49} & \multicolumn{1}{c}{0.52} & \multicolumn{1}{c}{0.39} & \multicolumn{1}{c|}{0.29} & \multicolumn{1}{c}{14.55} & \multicolumn{1}{c}{14.12} & \multicolumn{1}{c}{14.07} & \multicolumn{1}{c|}{11.23} & \multicolumn{1}{c}{1.76} & \multicolumn{1}{c}{1.73} & \multicolumn{1}{c}{1.50} & \multicolumn{1}{c|}{0.22} & \multicolumn{1}{c}{8.49} & \multicolumn{1}{c}{8.02} & \multicolumn{1}{c}{6.91} & \multicolumn{1}{c|}{2.99} & \multicolumn{1}{|c}{8.25} \\ \cmidrule(lr){3-3}
\multicolumn{1}{l|}{}                                                                                            & \multicolumn{1}{l|}{}                                                                                             & \multicolumn{1}{l|}{Z-score}                                                                                       & \multicolumn{1}{c}{22.27} & \multicolumn{1}{c}{21.50} & \multicolumn{1}{c}{20.56} & \multicolumn{1}{c|}{13.50} & \multicolumn{1}{c}{0.49} & \multicolumn{1}{c}{0.52} & \multicolumn{1}{c}{0.39} & \multicolumn{1}{c|}{0.29} & \multicolumn{1}{c}{14.55} & \multicolumn{1}{c}{14.12} & \multicolumn{1}{c}{14.07} & \multicolumn{1}{c|}{11.23} & \multicolumn{1}{c}{1.76} & \multicolumn{1}{c}{1.73} & \multicolumn{1}{c}{1.50} & \multicolumn{1}{c|}{0.22} & \multicolumn{1}{c}{8.49} & \multicolumn{1}{c}{8.02} & \multicolumn{1}{c}{6.91} & \multicolumn{1}{c|}{2.99} & \multicolumn{1}{|c}{8.25} \\ \cmidrule(lr){2-3}
\multicolumn{1}{l|}{}                                                                                            & \multicolumn{1}{l|}{\multirow{2}{*}{Uncertainty}}                                                                 & \multicolumn{1}{l|}{All-model}                                                                                     & \multicolumn{1}{c}{22.27} & \multicolumn{1}{c}{18.75} & \multicolumn{1}{c}{14.67} & \multicolumn{1}{c|}{0.04} & \multicolumn{1}{c}{0.49} & \multicolumn{1}{c}{0.00} & \multicolumn{1}{c}{0.00} & \multicolumn{1}{c|}{0.00} & \multicolumn{1}{c}{14.55} & \multicolumn{1}{c}{10.74} & \multicolumn{1}{c}{5.64} & \multicolumn{1}{c|}{1.26} & \multicolumn{1}{c}{1.76} & \multicolumn{1}{c}{0.14} & \multicolumn{1}{c}{0.11} & \multicolumn{1}{c|}{0.00} & \multicolumn{1}{c}{8.49} & \multicolumn{1}{c}{5.01} & \multicolumn{1}{c}{4.18} & \multicolumn{1}{c|}{1.24} & \multicolumn{1}{|c}{5.47} \\ \cmidrule(lr){3-3}
\multicolumn{1}{l|}{}                                                                                            & \multicolumn{1}{l|}{}                                                                                             & \multicolumn{1}{l|}{Z-score}                                                                                       & \multicolumn{1}{c}{22.27} & \multicolumn{1}{c}{17.64} & \multicolumn{1}{c}{12.75} & \multicolumn{1}{c|}{0.20} & \multicolumn{1}{c}{0.49} & \multicolumn{1}{c}{0.00} & \multicolumn{1}{c}{0.00} & \multicolumn{1}{c|}{0.00} & \multicolumn{1}{c}{14.55} & \multicolumn{1}{c}{11.22} & \multicolumn{1}{c}{5.43} & \multicolumn{1}{c|}{0.51} & \multicolumn{1}{c}{1.76} & \multicolumn{1}{c}{0.13} & \multicolumn{1}{c}{0.10} & \multicolumn{1}{c|}{0.00} & \multicolumn{1}{c}{8.49} & \multicolumn{1}{c}{4.63} & \multicolumn{1}{c}{2.94} & \multicolumn{1}{c|}{0.52} & \multicolumn{1}{|c}{5.18} \\ \cmidrule(lr){2-3}
\multicolumn{1}{l|}{}                                                                                            & \multicolumn{1}{l|}{\multirow{2}{*}{Margin}}                                                                      & \multicolumn{1}{l|}{All-model}                                                                                     & \multicolumn{1}{c}{22.27} & \multicolumn{1}{c}{18.75} & \multicolumn{1}{c}{14.67} & \multicolumn{1}{c|}{0.04} & \multicolumn{1}{c}{0.49} & \multicolumn{1}{c}{0.00} & \multicolumn{1}{c}{0.00} & \multicolumn{1}{c|}{0.00} & \multicolumn{1}{c}{14.55} & \multicolumn{1}{c}{10.74} & \multicolumn{1}{c}{5.64} & \multicolumn{1}{c|}{1.26} & \multicolumn{1}{c}{1.76} & \multicolumn{1}{c}{0.13} & \multicolumn{1}{c}{0.04} & \multicolumn{1}{c|}{0.00} & \multicolumn{1}{c}{8.49} & \multicolumn{1}{c}{5.01} & \multicolumn{1}{c}{4.15} & \multicolumn{1}{c|}{1.01} & \multicolumn{1}{|c}{5.45} \\ \cmidrule(lr){3-3} 
\multicolumn{1}{l|}{}                                                                                            & \multicolumn{1}{l|}{}                                                                                             & \multicolumn{1}{l|}{Z-score}                                                                                       & \multicolumn{1}{c}{22.27} & \multicolumn{1}{c}{17.64} & \multicolumn{1}{c}{12.75} & \multicolumn{1}{c|}{0.20} & \multicolumn{1}{c}{0.49} & \multicolumn{1}{c}{0.00} & \multicolumn{1}{c}{0.00} & \multicolumn{1}{c|}{0.00} & \multicolumn{1}{c}{14.55} & \multicolumn{1}{c}{11.22} & \multicolumn{1}{c}{5.43} & \multicolumn{1}{c|}{0.51} & \multicolumn{1}{c}{1.76} & \multicolumn{1}{c}{0.13} & \multicolumn{1}{c}{0.08} & \multicolumn{1}{c|}{0.00} & \multicolumn{1}{c}{8.49} & \multicolumn{1}{c}{4.38} & \multicolumn{1}{c}{2.89} & \multicolumn{1}{c|}{0.32} & \multicolumn{1}{|c}{5.16} \\\cmidrule(lr){2-3}
\multicolumn{1}{l|}{}                                                                                            & \multicolumn{1}{l|}{\multirow{2}{*}{Entropy}}                                                                     & \multicolumn{1}{l|}{All-model}                                                                                     & \multicolumn{1}{c}{22.27} & \multicolumn{1}{c}{18.75} & \multicolumn{1}{c}{14.67} & \multicolumn{1}{c|}{0.04} & \multicolumn{1}{c}{0.49} & \multicolumn{1}{c}{0.00} & \multicolumn{1}{c}{0.00} & \multicolumn{1}{c|}{0.00} & \multicolumn{1}{c}{14.55} & \multicolumn{1}{c}{10.74} & \multicolumn{1}{c}{5.64} & \multicolumn{1}{c|}{1.26} & \multicolumn{1}{c}{1.76} & \multicolumn{1}{c}{0.04} & \multicolumn{1}{c}{0.12} & \multicolumn{1}{c|}{0.00} & \multicolumn{1}{c}{8.49} & \multicolumn{1}{c}{5.62} & \multicolumn{1}{c}{4.68} & \multicolumn{1}{c|}{1.20} & \multicolumn{1}{|c}{5.52} \\ \cmidrule(lr){3-3}
\multicolumn{1}{l|}{}                                                                                            & \multicolumn{1}{l|}{}                                                                                             & \multicolumn{1}{l|}{Z-score}                                                                                       & \multicolumn{1}{c}{22.27} & \multicolumn{1}{c}{17.64} & \multicolumn{1}{c}{12.75} & \multicolumn{1}{c|}{0.20} & \multicolumn{1}{c}{0.49} & \multicolumn{1}{c}{0.00} & \multicolumn{1}{c}{0.00} & \multicolumn{1}{c|}{0.00} & \multicolumn{1}{c}{14.55} & \multicolumn{1}{c}{11.22} & \multicolumn{1}{c}{5.43} & \multicolumn{1}{c|}{0.51} & \multicolumn{1}{c}{1.76} & \multicolumn{1}{c}{0.09} & \multicolumn{1}{c}{0.14} & \multicolumn{1}{c|}{0.00} & \multicolumn{1}{c}{8.49} & \multicolumn{1}{c}{4.90} & \multicolumn{1}{c}{3.17} & \multicolumn{1}{c|}{0.66} & \multicolumn{1}{|c}{5.21} \\ \toprule
\multicolumn{1}{l|}{\multirow{10}{*}{Soft Ensemble}}                                                              & \multicolumn{1}{l|}{\multirow{2}{*}{Random}}                                                                      & \multicolumn{1}{l|}{All-model}                                                                                     & \multicolumn{1}{c}{22.16} & \multicolumn{1}{c}{21.17} & \multicolumn{1}{c}{20.09} & \multicolumn{1}{c|}{13.30} & \multicolumn{1}{c}{0.49} & \multicolumn{1}{c}{0.52} & \multicolumn{1}{c}{0.39} & \multicolumn{1}{c|}{0.29} & \multicolumn{1}{c}{14.19} & \multicolumn{1}{c}{13.67} & \multicolumn{1}{c}{13.54} & \multicolumn{1}{c|}{11.33} & \multicolumn{1}{c}{1.76} & \multicolumn{1}{c}{1.74} & \multicolumn{1}{c}{1.53} & \multicolumn{1}{c|}{0.24} & \multicolumn{1}{c}{8.86} & \multicolumn{1}{c}{8.69} & \multicolumn{1}{c}{7.79} & \multicolumn{1}{c|}{3.61} & \multicolumn{1}{|c}{8.27} \\ \cmidrule(lr){3-3}
\multicolumn{1}{l|}{}                                                                                            & \multicolumn{1}{l|}{}                                                                                             & \multicolumn{1}{l|}{Z-score}                                                                                       & \multicolumn{1}{c}{22.16} & \multicolumn{1}{c}{21.17} & \multicolumn{1}{c}{20.09} & \multicolumn{1}{c|}{13.30} & \multicolumn{1}{c}{0.49} & \multicolumn{1}{c}{0.52} & \multicolumn{1}{c}{0.39} & \multicolumn{1}{c|}{0.29} & \multicolumn{1}{c}{14.19} & \multicolumn{1}{c}{13.67} & \multicolumn{1}{c}{13.54} & \multicolumn{1}{c|}{11.33} & \multicolumn{1}{c}{1.76} & \multicolumn{1}{c}{1.74} & \multicolumn{1}{c}{1.53} & \multicolumn{1}{c|}{0.24} & \multicolumn{1}{c}{8.86} & \multicolumn{1}{c}{8.69} & \multicolumn{1}{c}{7.79} & \multicolumn{1}{c|}{3.61} & \multicolumn{1}{|c}{8.27} \\ \cmidrule(lr){2-3}
\multicolumn{1}{l|}{}                                                                                            & \multicolumn{1}{l|}{\multirow{2}{*}{Uncertainty}}                                                                 & \multicolumn{1}{l|}{All-model}                                                                                     & \multicolumn{1}{c}{22.16} & \multicolumn{1}{c}{18.24} & \multicolumn{1}{c}{13.69} & \multicolumn{1}{c|}{0.41} & \multicolumn{1}{c}{0.49} & \multicolumn{1}{c}{0.01} & \multicolumn{1}{c}{0.00} & \multicolumn{1}{c|}{0.00} & \multicolumn{1}{c}{14.19} & \multicolumn{1}{c}{10.21} & \multicolumn{1}{c}{5.53} & \multicolumn{1}{c|}{0.25} & \multicolumn{1}{c}{1.76} & \multicolumn{1}{c}{0.14} & \multicolumn{1}{c}{0.14} & \multicolumn{1}{c|}{0.00} & \multicolumn{1}{c}{8.86} & \multicolumn{1}{c}{4.98} & \multicolumn{1}{c}{4.10} & \multicolumn{1}{c|}{0.87} & \multicolumn{1}{|c}{5.30} \\ \cmidrule(lr){3-3}
\multicolumn{1}{l|}{}                                                                                            & \multicolumn{1}{l|}{}                                                                                             & \multicolumn{1}{l|}{Z-score}                                                                                       & \multicolumn{1}{c}{22.16} & \multicolumn{1}{c}{16.89} & \multicolumn{1}{c}{12.96} & \multicolumn{1}{c|}{0.07} & \multicolumn{1}{c}{0.49} & \multicolumn{1}{c}{0.01} & \multicolumn{1}{c}{0.00} & \multicolumn{1}{c|}{0.00} & \multicolumn{1}{c}{14.19} & \multicolumn{1}{c}{10.76} & \multicolumn{1}{c}{6.44} & \multicolumn{1}{c|}{0.63} & \multicolumn{1}{c}{1.76} & \multicolumn{1}{c}{0.14} & \multicolumn{1}{c}{0.14} & \multicolumn{1}{c|}{0.00} & \multicolumn{1}{c}{8.86} & \multicolumn{1}{c}{4.77} & \multicolumn{1}{c}{3.63} & \multicolumn{1}{c|}{0.85} & \multicolumn{1}{|c}{5.24} \\ \cmidrule(lr){2-3}
\multicolumn{1}{l|}{}                                                                                            & \multicolumn{1}{l|}{\multirow{2}{*}{Margin}}                                                                      & \multicolumn{1}{l|}{All-model}                                                                                     & \multicolumn{1}{c}{22.16} & \multicolumn{1}{c}{18.24} & \multicolumn{1}{c}{13.69} & \multicolumn{1}{c|}{0.41} & \multicolumn{1}{c}{0.49} & \multicolumn{1}{c}{0.01} & \multicolumn{1}{c}{0.00} & \multicolumn{1}{c|}{0.00} & \multicolumn{1}{c}{14.19} & \multicolumn{1}{c}{10.21} & \multicolumn{1}{c}{5.53} & \multicolumn{1}{c|}{0.25} & \multicolumn{1}{c}{1.76} & \multicolumn{1}{c}{0.05} & \multicolumn{1}{c}{0.14} & \multicolumn{1}{c|}{0.00} & \multicolumn{1}{c}{8.86} & \multicolumn{1}{c}{4.97} & \multicolumn{1}{c}{3.35} & \multicolumn{1}{c|}{1.18} & \multicolumn{1}{|c}{5.27} \\ \cmidrule(lr){3-3}
\multicolumn{1}{l|}{}                                                                                            & \multicolumn{1}{l|}{}                                                                                             & \multicolumn{1}{l|}{Z-score}                                                                                       & \multicolumn{1}{c}{22.16} & \multicolumn{1}{c}{16.89} & \multicolumn{1}{c}{12.96} & \multicolumn{1}{c|}{0.07} & \multicolumn{1}{c}{0.49} & \multicolumn{1}{c}{0.01} & \multicolumn{1}{c}{0.00} & \multicolumn{1}{c|}{0.00} & \multicolumn{1}{c}{14.19} & \multicolumn{1}{c}{10.76} & \multicolumn{1}{c}{6.44} & \multicolumn{1}{c|}{0.63} & \multicolumn{1}{c}{1.76} & \multicolumn{1}{c}{0.05} & \multicolumn{1}{c}{0.14} & \multicolumn{1}{c|}{0.00} & \multicolumn{1}{c}{8.86} & \multicolumn{1}{c}{4.77} & \multicolumn{1}{c}{2.90} & \multicolumn{1}{c|}{0.97} & \multicolumn{1}{|c}{5.20} \\ \cmidrule(lr){2-3}
\multicolumn{1}{l|}{}                                                                                            & \multicolumn{1}{l|}{\multirow{2}{*}{Entropy}}                                                                     & \multicolumn{1}{l|}{All-model}                                                                                     & \multicolumn{1}{c}{22.16} & \multicolumn{1}{c}{18.24} & \multicolumn{1}{c}{13.69} & \multicolumn{1}{c|}{0.41} & \multicolumn{1}{c}{0.49} & \multicolumn{1}{c}{0.01} & \multicolumn{1}{c}{0.00} & \multicolumn{1}{c|}{0.00} & \multicolumn{1}{c}{14.19} & \multicolumn{1}{c}{10.21} & \multicolumn{1}{c}{5.53} & \multicolumn{1}{c|}{0.25} & \multicolumn{1}{c}{1.76} & \multicolumn{1}{c}{0.06} & \multicolumn{1}{c}{0.10} & \multicolumn{1}{c|}{0.00} & \multicolumn{1}{c}{8.86} & \multicolumn{1}{c}{6.04} & \multicolumn{1}{c}{4.50} & \multicolumn{1}{c|}{1.07} & \multicolumn{1}{|c}{5.38} \\ \cmidrule(lr){3-3}
\multicolumn{1}{l|}{}                                                                                            & \multicolumn{1}{l|}{}                                                                                             & \multicolumn{1}{l|}{Z-score}                                                                                       & \multicolumn{1}{c}{22.16} & \multicolumn{1}{c}{16.89} & \multicolumn{1}{c}{12.96} & \multicolumn{1}{c|}{0.07} & \multicolumn{1}{c}{0.49} & \multicolumn{1}{c}{0.01} & \multicolumn{1}{c}{0.00} & \multicolumn{1}{c|}{0.00} & \multicolumn{1}{c}{14.19} & \multicolumn{1}{c}{10.76} & \multicolumn{1}{c}{6.44} & \multicolumn{1}{c|}{0.63} & \multicolumn{1}{c}{1.76} & \multicolumn{1}{c}{0.10} & \multicolumn{1}{c}{0.14} & \multicolumn{1}{c|}{0.00} & \multicolumn{1}{c}{8.86} & \multicolumn{1}{c}{5.00} & \multicolumn{1}{c}{3.73} & \multicolumn{1}{c|}{0.83} & \multicolumn{1}{|c}{5.25} \\
\bottomrule
\bottomrule
\end{tabular}
}
\caption{
\textbf{Weak supervision setting}: 
This table illustrates the changes in optimal gap values within our design space.
These changes are observed across different budget ratios, specifically at 0\%, 10\%, 20\%, and 50\%.
}
\label{tab:wrench_og_diff_ratio}
\end{table*}

\begin{figure*}[t]
\centering
\includegraphics[scale=0.165]{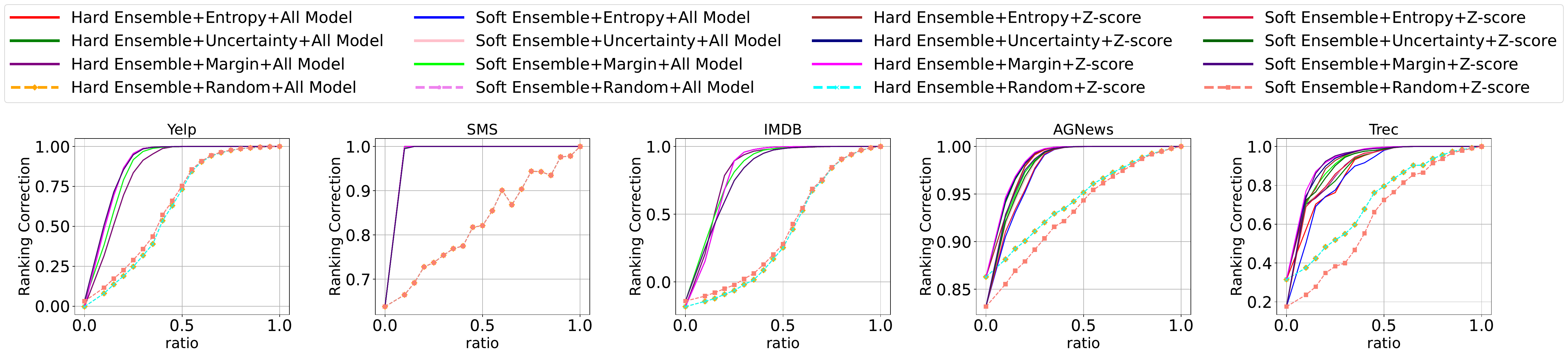}
\caption{\textbf{Weak supervision setting}: This figure illustrates the changes in ranking correction values within our design space. 
These changes are observed across budget ratios from 0 to 1.}
\vspace{-1ex}
\label{fig:wrench_rc}
\end{figure*}

\begin{table*}[t]
\centering
\scalebox{0.43}{

\begin{tabular}{@{}lll|lllllllllllllllllllllllll@{}}
\toprule
\bottomrule
\multicolumn{3}{c|}{\textbf{Method}}                                                                                                                                                                                                                                                                                                                    & \multicolumn{24}{c}{\textbf{Dataset}}                                                                                                                                                                                        &       \\ \midrule
\multicolumn{1}{c}{\multirow{2}{*}{\textbf{\begin{tabular}[c]{@{}c@{}}Pseudo-label\\ Generation\end{tabular}}}} & \multicolumn{1}{c}{\multirow{2}{*}{\textbf{\begin{tabular}[c]{@{}c@{}}Active Label\\ Acquisition\end{tabular}}}} & \multicolumn{1}{c|}{\multirow{2}{*}{\textbf{\begin{tabular}[c]{@{}c@{}}Model Committee\\ Selection\end{tabular}}}} & \multicolumn{3}{c|}{\textbf{WSC}} & \multicolumn{3}{c|}{\textbf{Story}} & \multicolumn{3}{c|}{\textbf{CB}} & \multicolumn{3}{c|}{\textbf{RTE}} & \multicolumn{3}{c|}{\textbf{WiC}} & \multicolumn{3}{c|}{\textbf{ANLI1}} & \multicolumn{3}{c|}{\textbf{ANLI2}} & \multicolumn{3}{c|}{\textbf{ANLI3}} & \textbf{Avg.}  \\ \cmidrule(lr){4-28}
\multicolumn{1}{c}{}                                                                                            & \multicolumn{1}{c}{}                                                                                             & \multicolumn{1}{c|}{}                                                                                              & 0\%   & 10\%  & \multicolumn{1}{c|}{30\%} & 0\%   & 10\%  & \multicolumn{1}{c|}{30\%} & 0\%   & 10\%  & \multicolumn{1}{c|}{30\%} & 0\%   & 10\%  & \multicolumn{1}{c|}{30\%} & 0\%   & 10\%  & \multicolumn{1}{c|}{30\%} & 0\%   & 10\%  & \multicolumn{1}{c|}{30\%} & 0\%   & 10\%  & \multicolumn{1}{c|}{30\%} & 0\%   & 10\%  & \multicolumn{1}{c|}{30\%} &       \\ \toprule
\multicolumn{1}{l|}{\multirow{10}{*}{Hard Ensemble}}                                                             & \multicolumn{1}{l|}{\multirow{2}{*}{Random}}                                                                     & All Model                                                                                                          & \multicolumn{1}{c}{1.16} & \multicolumn{1}{c}{0.95} & \multicolumn{1}{c|}{1.04} & \multicolumn{1}{c}{0.03} & \multicolumn{1}{c}{0.02} & \multicolumn{1}{c|}{0.01} & \multicolumn{1}{c}{2.84} & \multicolumn{1}{c}{2.67} & \multicolumn{1}{c|}{1.82} & \multicolumn{1}{c}{0.40} & \multicolumn{1}{c}{0.38} & \multicolumn{1}{c|}{0.50} & \multicolumn{1}{c}{1.14} & \multicolumn{1}{c}{0.86} & \multicolumn{1}{c|}{0.82} & \multicolumn{1}{c}{0.05} & \multicolumn{1}{c}{0.06} & \multicolumn{1}{c|}{0.06} & \multicolumn{1}{c}{0.46} & \multicolumn{1}{c}{0.48} & \multicolumn{1}{c|}{0.40} & \multicolumn{1}{c}{0.81} & \multicolumn{1}{c}{0.80} & \multicolumn{1}{c}{0.82} & \multicolumn{1}{|c}{0.77} \\ \cmidrule(lr){3-3}
\multicolumn{1}{l|}{}                                                                                           & \multicolumn{1}{l|}{}                                                                                            & Z-score                                                                                                             & \multicolumn{1}{c}{1.16} & \multicolumn{1}{c}{0.95} & \multicolumn{1}{c|}{1.04} & \multicolumn{1}{c}{0.03} & \multicolumn{1}{c}{0.02} & \multicolumn{1}{c|}{0.01} & \multicolumn{1}{c}{2.84} & \multicolumn{1}{c}{2.67} & \multicolumn{1}{c|}{1.82} & \multicolumn{1}{c}{0.40} & \multicolumn{1}{c}{0.38} & \multicolumn{1}{c|}{0.50} & \multicolumn{1}{c}{1.14} & \multicolumn{1}{c}{0.86} & \multicolumn{1}{c|}{0.82} & \multicolumn{1}{c}{0.05} & \multicolumn{1}{c}{0.06} & \multicolumn{1}{c|}{0.06} & \multicolumn{1}{c}{0.46} & \multicolumn{1}{c}{0.48} & \multicolumn{1}{c|}{0.40} & \multicolumn{1}{c}{0.81} & \multicolumn{1}{c}{0.80} & \multicolumn{1}{c}{0.82} & \multicolumn{1}{|c}{0.77} \\ \cmidrule(lr){2-3}
\multicolumn{1}{l|}{}                                                                                           & \multicolumn{1}{l|}{\multirow{2}{*}{Uncertainty}}                                                                & All Model                                                                                                          & \multicolumn{1}{c}{1.16} & \multicolumn{1}{c}{0.64} & \multicolumn{1}{c|}{0.03} & \multicolumn{1}{c}{0.03} & \multicolumn{1}{c}{0.03} & \multicolumn{1}{c|}{0.01} & \multicolumn{1}{c}{2.84} & \multicolumn{1}{c}{1.60} & \multicolumn{1}{c|}{0.40} & \multicolumn{1}{c}{0.40} & \multicolumn{1}{c}{0.07} & \multicolumn{1}{c|}{0.40} & \multicolumn{1}{c}{1.14} & \multicolumn{1}{c}{1.05} & \multicolumn{1}{c|}{0.04} & \multicolumn{1}{c}{0.05} & \multicolumn{1}{c}{0.07} & \multicolumn{1}{c|}{0.46} & \multicolumn{1}{c}{0.46} & \multicolumn{1}{c}{0.33} & \multicolumn{1}{c|}{0.44} & \multicolumn{1}{c}{0.81} & \multicolumn{1}{c}{0.86} & \multicolumn{1}{c}{0.85} & \multicolumn{1}{|c}{0.59} \\ \cmidrule(lr){3-3}
\multicolumn{1}{l|}{}                                                                                           & \multicolumn{1}{l|}{}                                                                                            & Z-score                                                                                                            & \multicolumn{1}{c}{1.16} & \multicolumn{1}{c}{0.00} & \multicolumn{1}{c|}{0.03} & \multicolumn{1}{c}{0.03} & \multicolumn{1}{c}{0.00} & \multicolumn{1}{c|}{0.00} & \multicolumn{1}{c}{2.84} & \multicolumn{1}{c}{0.00} & \multicolumn{1}{c|}{0.00} & \multicolumn{1}{c}{0.40} & \multicolumn{1}{c}{0.00} & \multicolumn{1}{c|}{0.00} & \multicolumn{1}{c}{1.14} & \multicolumn{1}{c}{1.19} & \multicolumn{1}{c|}{0.17} & \multicolumn{1}{c}{0.05} & \multicolumn{1}{c}{0.13} & \multicolumn{1}{c|}{0.57} & \multicolumn{1}{c}{0.46} & \multicolumn{1}{c}{0.26} & \multicolumn{1}{c|}{0.45} & \multicolumn{1}{c}{0.81} & \multicolumn{1}{c}{0.81} & \multicolumn{1}{c}{0.83} & \multicolumn{1}{|c}{0.47} \\ \cmidrule(lr){2-3}
\multicolumn{1}{l|}{}                                                                                           & \multicolumn{1}{l|}{\multirow{2}{*}{Margin}}                                                                     & All Model                                                                                                          & \multicolumn{1}{c}{1.16} & \multicolumn{1}{c}{0.64} & \multicolumn{1}{c|}{0.03} & \multicolumn{1}{c}{0.03} & \multicolumn{1}{c}{0.03} & \multicolumn{1}{c|}{0.01} & \multicolumn{1}{c}{2.84} & \multicolumn{1}{c}{1.64} & \multicolumn{1}{c|}{0.27} & \multicolumn{1}{c}{0.40} & \multicolumn{1}{c}{0.07} & \multicolumn{1}{c|}{0.40} & \multicolumn{1}{c}{1.14} & \multicolumn{1}{c}{1.05} & \multicolumn{1}{c|}{0.04} & \multicolumn{1}{c}{0.05} & \multicolumn{1}{c}{0.23} & \multicolumn{1}{c|}{0.55} & \multicolumn{1}{c}{0.46} & \multicolumn{1}{c}{0.33} & \multicolumn{1}{c|}{0.49} & \multicolumn{1}{c}{0.81} & \multicolumn{1}{c}{0.94} & \multicolumn{1}{c}{0.85} & \multicolumn{1}{|c}{0.60} \\ \cmidrule(lr){3-3}
\multicolumn{1}{l|}{}                                                                                           & \multicolumn{1}{l|}{}                                                                                            & Z-score                                                                                                            & \multicolumn{1}{c}{1.16} & \multicolumn{1}{c}{0.00} & \multicolumn{1}{c|}{0.03} & \multicolumn{1}{c}{0.03} & \multicolumn{1}{c}{0.00} & \multicolumn{1}{c|}{0.00} & \multicolumn{1}{c}{2.84} & \multicolumn{1}{c}{0.00} & \multicolumn{1}{c|}{0.00} & \multicolumn{1}{c}{0.40} & \multicolumn{1}{c}{0.00} & \multicolumn{1}{c|}{0.00} & \multicolumn{1}{c}{1.14} & \multicolumn{1}{c}{1.19} & \multicolumn{1}{c|}{0.17} & \multicolumn{1}{c}{0.05} & \multicolumn{1}{c}{0.11} & \multicolumn{1}{c|}{0.51} & \multicolumn{1}{c}{0.46} & \multicolumn{1}{c}{0.27} & \multicolumn{1}{c|}{0.42} & \multicolumn{1}{c}{0.81} & \multicolumn{1}{c}{0.77} & \multicolumn{1}{c}{0.75} & \multicolumn{1}{|c}{0.46} \\  \cmidrule(lr){2-3}
\multicolumn{1}{l|}{}                                                                                           & \multicolumn{1}{l|}{\multirow{2}{*}{Entropy}}                                                                    & All Model                                                                                                          & \multicolumn{1}{c}{1.16} & \multicolumn{1}{c}{0.64} & \multicolumn{1}{c|}{0.03} & \multicolumn{1}{c}{0.03} & \multicolumn{1}{c}{0.03} & \multicolumn{1}{c|}{0.01} & \multicolumn{1}{c}{2.84} & \multicolumn{1}{c}{1.42} & \multicolumn{1}{c|}{0.31} & \multicolumn{1}{c}{0.40} & \multicolumn{1}{c}{0.07} & \multicolumn{1}{c|}{0.40} & \multicolumn{1}{c}{1.14} & \multicolumn{1}{c}{1.05} & \multicolumn{1}{c|}{0.04} & \multicolumn{1}{c}{0.05} & \multicolumn{1}{c}{0.10} & \multicolumn{1}{c|}{0.51} & \multicolumn{1}{c}{0.46} & \multicolumn{1}{c}{0.31} & \multicolumn{1}{c|}{0.45} & \multicolumn{1}{c}{0.81} & \multicolumn{1}{c}{0.66} & \multicolumn{1}{c}{0.85} & \multicolumn{1}{|c}{0.57} \\ \cmidrule(lr){3-3}
\multicolumn{1}{l|}{}                                                                                           & \multicolumn{1}{l|}{}                                                                                            & Z-score                                                                                                            & \multicolumn{1}{c}{1.16} & \multicolumn{1}{c}{0.00} & \multicolumn{1}{c|}{0.03} & \multicolumn{1}{c}{0.03} & \multicolumn{1}{c}{0.00} & \multicolumn{1}{c|}{0.00} & \multicolumn{1}{c}{2.84} & \multicolumn{1}{c}{0.00} & \multicolumn{1}{c|}{0.00} & \multicolumn{1}{c}{0.40} & \multicolumn{1}{c}{0.00} & \multicolumn{1}{c|}{0.00} & \multicolumn{1}{c}{1.14} & \multicolumn{1}{c}{1.19} & \multicolumn{1}{c|}{0.17} & \multicolumn{1}{c}{0.05} & \multicolumn{1}{c}{0.08} & \multicolumn{1}{c|}{0.56} & \multicolumn{1}{c}{0.46} & \multicolumn{1}{c}{0.31} & \multicolumn{1}{c|}{0.43} & \multicolumn{1}{c}{0.81} & \multicolumn{1}{c}{0.83} & \multicolumn{1}{c}{0.76} & \multicolumn{1}{|c}{0.47} \\ \toprule
\multicolumn{1}{l|}{\multirow{10}{*}{Soft Ensemble}}                                                             & \multicolumn{1}{l|}{\multirow{2}{*}{Random}}                                                                     & All Model                                                                                                          & \multicolumn{1}{c}{2.12} & \multicolumn{1}{c}{2.01} & \multicolumn{1}{c|}{1.33} & \multicolumn{1}{c}{0.04} & \multicolumn{1}{c}{0.04} & \multicolumn{1}{c|}{0.02} & \multicolumn{1}{c}{2.84} & \multicolumn{1}{c}{2.71} & \multicolumn{1}{c|}{1.91} & \multicolumn{1}{c}{1.40} & \multicolumn{1}{c}{1.33} & \multicolumn{1}{c|}{1.02} & \multicolumn{1}{c}{0.19} & \multicolumn{1}{c}{0.10} & \multicolumn{1}{c|}{0.16} & \multicolumn{1}{c}{0.20} & \multicolumn{1}{c}{0.22} & \multicolumn{1}{c|}{0.08} & \multicolumn{1}{c}{0.55} & \multicolumn{1}{c}{0.54} & \multicolumn{1}{c|}{0.57} & \multicolumn{1}{c}{1.18} & \multicolumn{1}{c}{1.17} & \multicolumn{1}{c}{1.13} & \multicolumn{1}{|c}{0.95} \\ \cmidrule(lr){3-3}
\multicolumn{1}{l|}{}                                                                                           & \multicolumn{1}{l|}{}                                                                                            & Z-score                                                                                                            & \multicolumn{1}{c}{2.12} & \multicolumn{1}{c}{2.01} & \multicolumn{1}{c|}{1.33} & \multicolumn{1}{c}{0.04} & \multicolumn{1}{c}{0.04} & \multicolumn{1}{c|}{0.02} & \multicolumn{1}{c}{2.84} & \multicolumn{1}{c}{2.71} & \multicolumn{1}{c|}{1.91} & \multicolumn{1}{c}{1.40} & \multicolumn{1}{c}{1.33} & \multicolumn{1}{c|}{1.02} & \multicolumn{1}{c}{0.19} & \multicolumn{1}{c}{0.10} & \multicolumn{1}{c|}{0.16} & \multicolumn{1}{c}{0.20} & \multicolumn{1}{c}{0.22} & \multicolumn{1}{c|}{0.08} & \multicolumn{1}{c}{0.55} & \multicolumn{1}{c}{0.54} & \multicolumn{1}{c|}{0.57} & \multicolumn{1}{c}{1.18} & \multicolumn{1}{c}{1.17} & \multicolumn{1}{c}{1.13} & \multicolumn{1}{|c}{0.95} \\ \cmidrule(lr){2-3}
\multicolumn{1}{l|}{}                                                                                           & \multicolumn{1}{l|}{\multirow{2}{*}{Uncertainty}}                                                                & All Model                                                                                                          & \multicolumn{1}{c}{2.12} & \multicolumn{1}{c}{1.10} & \multicolumn{1}{c|}{0.04} & \multicolumn{1}{c}{0.04} & \multicolumn{1}{c}{0.01} & \multicolumn{1}{c|}{0.00} & \multicolumn{1}{c}{2.84} & \multicolumn{1}{c}{1.29} & \multicolumn{1}{c|}{0.36} & \multicolumn{1}{c}{1.40} & \multicolumn{1}{c}{2.64} & \multicolumn{1}{c|}{2.00} & \multicolumn{1}{c}{0.19} & \multicolumn{1}{c}{0.52} & \multicolumn{1}{c|}{0.14} & \multicolumn{1}{c}{0.20} & \multicolumn{1}{c}{0.22} & \multicolumn{1}{c|}{0.58} & \multicolumn{1}{c}{0.55} & \multicolumn{1}{c}{0.44} & \multicolumn{1}{c|}{0.37} & \multicolumn{1}{c}{1.18} & \multicolumn{1}{c}{0.99} & \multicolumn{1}{c}{0.80} & \multicolumn{1}{|c}{0.83} \\ \cmidrule(lr){3-3}
\multicolumn{1}{l|}{}                                                                                           & \multicolumn{1}{l|}{}                                                                                            & Z-score                                                                                                            & \multicolumn{1}{c}{2.12} & \multicolumn{1}{c}{0.00} & \multicolumn{1}{c|}{0.04} & \multicolumn{1}{c}{0.04} & \multicolumn{1}{c}{0.00} & \multicolumn{1}{c|}{0.00} & \multicolumn{1}{c}{2.84} & \multicolumn{1}{c}{0.00} & \multicolumn{1}{c|}{0.00} & \multicolumn{1}{c}{1.40} & \multicolumn{1}{c}{0.00} & \multicolumn{1}{c|}{0.00} & \multicolumn{1}{c}{0.19} & \multicolumn{1}{c}{0.77} & \multicolumn{1}{c|}{0.16} & \multicolumn{1}{c}{0.20} & \multicolumn{1}{c}{0.20} & \multicolumn{1}{c|}{0.43} & \multicolumn{1}{c}{0.55} & \multicolumn{1}{c}{0.47} & \multicolumn{1}{c|}{0.33} & \multicolumn{1}{c}{1.18} & \multicolumn{1}{c}{0.88} & \multicolumn{1}{c}{0.97} & \multicolumn{1}{|c}{0.53} \\ \cmidrule(lr){2-3}
\multicolumn{1}{l|}{}                                                                                           & \multicolumn{1}{l|}{\multirow{2}{*}{Margin}}                                                                     & All Model                                                                                                          & \multicolumn{1}{c}{2.12} & \multicolumn{1}{c}{1.10} & \multicolumn{1}{c|}{0.04} & \multicolumn{1}{c}{0.04} & \multicolumn{1}{c}{0.01} & \multicolumn{1}{c|}{0.00} & \multicolumn{1}{c}{2.84} & \multicolumn{1}{c}{1.20} & \multicolumn{1}{c|}{0.04} & \multicolumn{1}{c}{1.40} & \multicolumn{1}{c}{2.64} & \multicolumn{1}{c|}{2.00} & \multicolumn{1}{c}{0.19} & \multicolumn{1}{c}{0.52} & \multicolumn{1}{c|}{0.14} & \multicolumn{1}{c}{0.20} & \multicolumn{1}{c}{0.36} & \multicolumn{1}{c|}{0.55} & \multicolumn{1}{c}{0.55} & \multicolumn{1}{c}{0.50} & \multicolumn{1}{c|}{0.31} & \multicolumn{1}{c}{1.18} & \multicolumn{1}{c}{1.07} & \multicolumn{1}{c}{0.90} & \multicolumn{1}{|c}{0.83} \\ \cmidrule(lr){3-3}
\multicolumn{1}{l|}{}                                                                                           & \multicolumn{1}{l|}{}                                                                                            & Z-score                                                                                                            & \multicolumn{1}{c}{2.12} & \multicolumn{1}{c}{0.00} & \multicolumn{1}{c|}{0.04} & \multicolumn{1}{c}{0.04} & \multicolumn{1}{c}{0.00} & \multicolumn{1}{c|}{0.00} & \multicolumn{1}{c}{2.84} & \multicolumn{1}{c}{0.00} & \multicolumn{1}{c|}{0.00} & \multicolumn{1}{c}{1.40} & \multicolumn{1}{c}{0.00} & \multicolumn{1}{c|}{0.00} & \multicolumn{1}{c}{0.19} & \multicolumn{1}{c}{0.77} & \multicolumn{1}{c|}{0.16} & \multicolumn{1}{c}{0.20} & \multicolumn{1}{c}{0.32} & \multicolumn{1}{c|}{0.56} & \multicolumn{1}{c}{0.55} & \multicolumn{1}{c}{0.48} & \multicolumn{1}{c|}{0.41} & \multicolumn{1}{c}{1.18} & \multicolumn{1}{c}{1.04} & \multicolumn{1}{c}{0.91} & \multicolumn{1}{|c}{0.55} \\ \cmidrule(lr){2-3}
\multicolumn{1}{l|}{}                                                                                           & \multicolumn{1}{l|}{\multirow{2}{*}{Entropy}}                                                                    & All Model                                                                                                          & \multicolumn{1}{c}{2.12} & \multicolumn{1}{c}{1.10} & \multicolumn{1}{c|}{0.04} & \multicolumn{1}{c}{0.04} & \multicolumn{1}{c}{0.01} & \multicolumn{1}{c|}{0.00} & \multicolumn{1}{c}{2.84} & \multicolumn{1}{c}{1.42} & \multicolumn{1}{c|}{1.07} & \multicolumn{1}{c}{1.40} & \multicolumn{1}{c}{2.64} & \multicolumn{1}{c|}{2.00} & \multicolumn{1}{c}{0.19} & \multicolumn{1}{c}{0.52} & \multicolumn{1}{c|}{0.14} & \multicolumn{1}{c}{0.20} & \multicolumn{1}{c}{0.03} & \multicolumn{1}{c|}{0.30} & \multicolumn{1}{c}{0.55} & \multicolumn{1}{c}{0.44} & \multicolumn{1}{c|}{0.59} & \multicolumn{1}{c}{1.18} & \multicolumn{1}{c}{0.94} & \multicolumn{1}{c}{1.02} & \multicolumn{1}{|c}{0.87} \\\cmidrule(lr){3-3}
\multicolumn{1}{l|}{}                                                                                           & \multicolumn{1}{l|}{}                                                                                            & Z-score                                                                                                            & \multicolumn{1}{c}{2.12} & \multicolumn{1}{c}{0.00} & \multicolumn{1}{c|}{0.04} & \multicolumn{1}{c}{0.04} & \multicolumn{1}{c}{0.00} & \multicolumn{1}{c|}{0.00} & \multicolumn{1}{c}{2.84} & \multicolumn{1}{c}{0.00} & \multicolumn{1}{c|}{0.00} & \multicolumn{1}{c}{1.40} & \multicolumn{1}{c}{0.00} & \multicolumn{1}{c|}{0.00} & \multicolumn{1}{c}{0.19} & \multicolumn{1}{c}{0.77} & \multicolumn{1}{c|}{0.16} & \multicolumn{1}{c}{0.20} & \multicolumn{1}{c}{0.02} & \multicolumn{1}{c|}{0.27} & \multicolumn{1}{c}{0.55} & \multicolumn{1}{c}{0.44} & \multicolumn{1}{c|}{0.27} & \multicolumn{1}{c}{1.18} & \multicolumn{1}{c}{0.81} & \multicolumn{1}{c}{1.07} & \multicolumn{1}{|c}{0.52} \\ \bottomrule
\bottomrule
\end{tabular}

}
\caption{
\textbf{Prompt selection setting}: 
This table illustrates the changes in optimal gap values within our design space. 
These changes are observed across different budget ratios, specifically at 0\%, 10\%, and 30\%.
}
\label{tab:prompt_og_diff_ratio_t2}
\end{table*}

\begin{figure*}[t]
\centering
\includegraphics[scale=0.175]{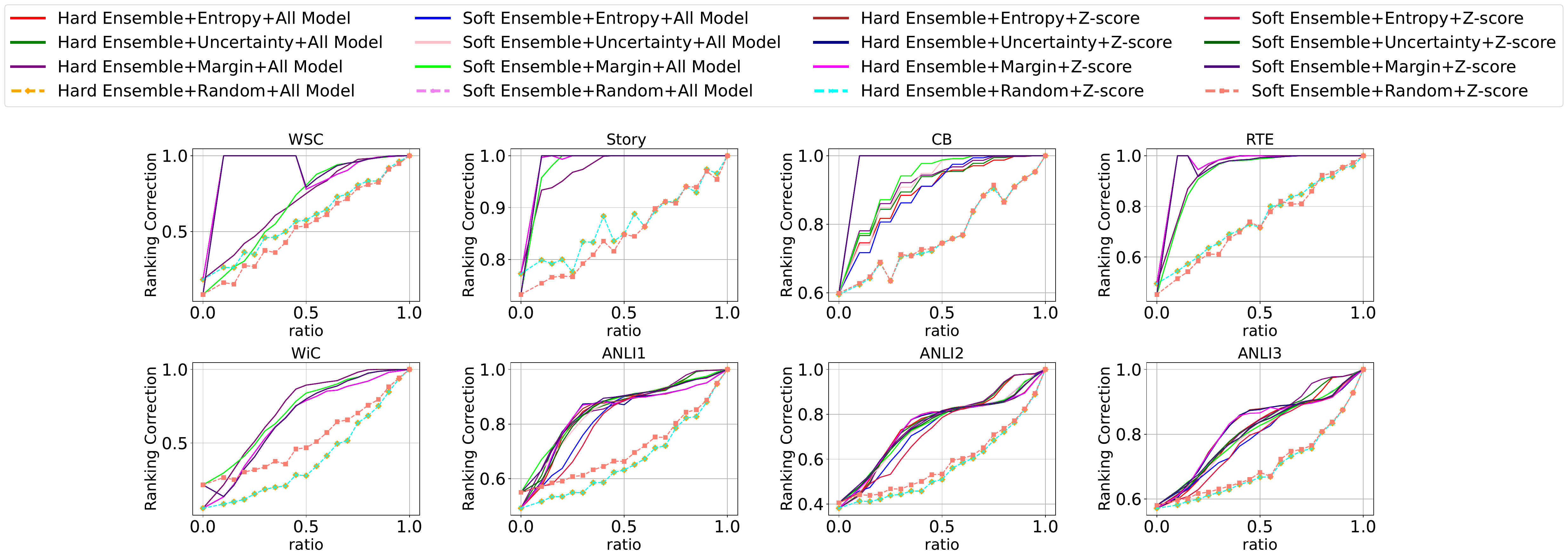}
\caption{\textbf{Prompt selection setting}: 
This figure illustrates the changes in ranking correction values within our design space. 
These changes are observed across budget ratios from 0 to 1.
The number after the dataset indicates the number of labels under the semi-supervised learning setting.
}
\vspace{-1ex}
\label{fig:prompt_rc}
\end{figure*}

\subsection{Step-IV: Model Ranking}
\label{sec:mr}

Step-IV in the \OURS{} framework is dedicated to ranking the models in the set $\mathcal{M}$. This step utilizes the final pseudo-label set $L_p$ and the ground-truth label set $L_g$ to evaluate each model's accuracy. The rank $r_p$ is determined as:
\begin{equation}
\begin{aligned}
r_{p} \leftarrow r (L_p,L_g , \mathcal{M}),
\end{aligned}
\end{equation}
where $r (\cdot)$ is the ranking function. 
It ranks the models in $\mathcal{M}$ according to their accuracy on $L_p$ and $L_g$.

\section{The MoraBench Benchmark}
\label{sec:MoraBench}

To advance research in model ranking and evaluate various design choices in our \OURS{} framework, we introduce MoraBench (Model Ranking Benchmark). This benchmark comprises a collection of model outputs generated under diverse scenarios.
The description of all model sets within MoraBench and its generation configuration are given in Appendix~\ref{sec:modelset-gen}.
We then perform model selection based on these outputs.
MoraBench and its associated resources are available at \url{http://github.com/ppsmk388/MoraBench}.

\subsection{Evaluation Metrics}
\label{e-metric}

We define two metrics used to evaluate the effectiveness of model selection results, namely Optimal Gap and Ranking Correction.

\paragraph{Optimal Gap.}

The Optimal Gap is defined as the difference in test accuracy between the best model chosen by the fully-labeled validation set, and the best model identified by the methods to be assessed.

\paragraph{Ranking Correction.}

Ranking Correction measures the similarity between the model rankings based on the fully-labeled validation set and those obtained by methods to be assessed. This similarity is assessed using the Spearman rank correlation coefficient\footnote{\url{http://docs.scipy.org/doc/scipy/reference/generated/scipy.stats.spearmanr.html}}
, a common non-parametric method evaluating the monotonic relationship between two ranked variables.

\section{Experiments}

We test our \OURS{} with MoraBench in detail under three scenarios, i.e., semi-supervised learning (Section~\ref{exp:semi}), weak supervision (Section~\ref{exp:weak}),  and prompt selection (Section~\ref{exp:prompt}).
Corresponding implementation details and design space are described in Appendix~\ref{sec:exsetup}.

\subsection{Semi-supervised Learning Setting}
\label{exp:semi}

Here, we evaluate the \OURS{} framework under a semi-supervised learning setting. 
For clarity, we first introduce the concept of 'budget ratio', defined as the proportion of our budget relative to the size of the complete unlabeled validation set. 
We examined the performance of \OURS{} at different budget ratios (0\%, 10\%, 20\% and 50\%), and relevant results are detailed in Tables \ref{tab:usb_og_diff_ratio_t1} and \ref{tab:usb_og_diff_ratio_t2}.
The impact of varying budget ratios on ranking correction is shown in Figure~\ref{fig:usb_rc}. 
Additionally, Table~\ref{tab:usb_optimal_gap} highlights the minimum budget needed to achieve an optimal gap of 0.
The number in brackets after the dataset indicates the number of labels used in the model training stage.
The model set generation setups can be found in Appendix~\ref{sec:semi-setting}.

From the results, we have the following findings: 
\textbf{First}, \OURS{} significantly minimizes labeling costs for model selection. 
For instance, in setting AGNews (200), we only need to label 387 samples to select the same model as labeling the entire validation set of 2000 samples (see Table~\ref{tab:usb_optimal_gap}). 
\textbf{Second}, our results consistently show the superiority of uncertainty sampling over random sampling. Table~\ref{tab:usb_optimal_gap} illustrates that random sampling typically requires a significantly larger budget compared to uncertainty sampling. This trend is evident in Table~\ref{tab:usb_og_diff_ratio_t1} and Table~\ref{tab:usb_og_diff_ratio_t2} as well. Additionally, the curves representing the random strategy in Figure~\ref{fig:usb_rc} consistently lie below of other uncertainty sampling strategies.
\textbf{Finally}, the model committee selected by Z-score is better than All-model under our design space.
For example, in Table~\ref{tab:usb_og_diff_ratio_t1} and~\ref{tab:usb_og_diff_ratio_t2}, the Z-score has a smaller optimal gap than All-model in all cases.

\subsection{Weak Supervision Setting}
\label{exp:weak}

In this section, we employed the WRENCH~\cite{zhang2021wrench} to evaluate our \OURS{} framework within a weak supervision setting. 
we first evaluate \OURS{} in a low-budget setting.
Specifically, we test our framework with budget ratios of 0\%, 10\%, 20\%, and 50\%. The corresponding optimal values are displayed in Table~\ref{tab:wrench_og_diff_ratio}. 
Additionally, Figure~\ref{fig:usb_rc} illustrates the variation in ranking correction as the budget ratio increases from 0 to 1. 
The model set generation setups can be found in Appendix~\ref{sec:weak-setting}.
Some interesting observations are shown as follows:
\textbf{First}, \OURS{}, when combined with an appropriate selection of methods, significantly lowers the labeling cost for validation sets: As shown in Table \ref{tab:wrench_og_diff_ratio}, only 10\% of the labeling cost suffices to select the same model that would be chosen with a fully labeled validation set.
\textbf{Then}, compared to random sampling, uncertainty sampling strategies consistently exhibit superior performance. This is evident in Table~\ref{tab:wrench_og_diff_ratio}, where the optimal gap for random sampling is highest across all budgets.
Moreover, from Figure~\ref{fig:usb_rc}, we notice the random strategy has a curve below all uncertainty sampling strategies, which further supports our conclusion. 
\textbf{Finally}, adopting the Z-score method generally reduces labeling costs as evidenced by the lower optimal gap values in Table \ref{tab:wrench_og_diff_ratio}. 
This suggests that removing the model that contains noise helps to reduce the labeling cost.

\subsection{Prompt Selection Setting}
\label{exp:prompt}

In this section, we employ the T0 benchmark~\cite{sanh2021multitask} to test \OURS{} under the prompt selection task.
With a large language model, denoted as $M$, and a set of prompts $\{p_{k}\}_{k \in [K]}$, we can analogize $M (p_k)$ to $m_k$ and refer to Step-I (Section~\ref{sec:plg}) to Step-IV  (Section~\ref{sec:mr}) to get the model rank.
The experimental results, including the optimal gap for budget ratios of 0\%, 10\%, and 30\%, are summarized in Table~\ref{tab:prompt_og_diff_ratio_t2}. 
Additionally, Figure~\ref{fig:prompt_rc} visually represents the changes in ranking correction as budget ratios vary from 0 to 1. 
The setups for model set generation can be found in Appendix~\ref{sec:prompt-setting}.

\textbf{First}, in Figure~\ref{fig:prompt_rc}, we find that under a limited budget, soft ensemble yields higher quality model rank if the model in the model set performs poorly, whereas hard ensemble is the superior solution.
For example, in the low-budget case, hard ensemble is a better choice in tasks RTE, Story, and WSC, where models generally perform better. 
While in tasks WIC, ANL1, ANL2, and ANL3, where models perform worse, soft ensemble works better.
A similar situation can be found in the Yelp (250), Amazon (100), Amazon (250), Yahoo (500), and Yahoo (2000) datasets in the semi-supervised setting (in Figure~\ref{fig:usb_rc}) as well as in the AGNews dataset and Trec dataset in the weakly supervised setting (in Figure~\ref{fig:wrench_rc}).
An intuitive explanation is that 
when the model's performance in the model set is poor, soft ensemble can utilize all the model's uncertainty information about the data, while hard ensemble may rely too much on some wrong prediction results, so soft ensemble will be more suitable in this case. 
When the model's performance in the model set is relatively high, hard ensemble can filter out the noisy information, which is more conducive to obtaining a high quality rank.
When the models in the model set all perform exceptionally well (SMS task of weak supervision setting in Figure~\ref{fig:wrench_rc}) or when the model predictions in the model set are relatively consistent (CB tasks of prompt selection), the results of the hard ensemble and the soft ensemble will remain consistent.
\textbf{Moreover}, our framework exhibits a substantial reduction in the labeling costs for validation sets. For example, as demonstrated in Table~\ref{tab:prompt_og_diff_ratio_t2}, for the SMS task, achieving an optimal gap value of 0 necessitates only 10\%  budget ratio. 
\textbf{Besides}, we find that when the sampling strategy is random, the optimal gap of the random strategy is the largest regardless of the budget ratio in Table~\ref{tab:prompt_og_diff_ratio_t2}.
\textbf{Lastly}, we observe that using Z-score reduces the budget required for all tasks. On average, the Z-score method yields a lower optimal gap value in Table~\ref{tab:prompt_og_diff_ratio_t2}. 
This suggests that a high-quality committee can generate a high-quality model ranking.

\section{Conclusion}
In this paper, we introduce \OURS{}, a novel framework that significantly reduces labeling costs for model selection tasks, particularly under resource-limited settings. 
To evaluate \OURS{}, we propose the MoraBench Benchmark, a comprehensive collection of model outputs across diverse scenarios.
Demonstrated across 23 tasks, including semi-supervised learning, weak supervision, and prompt selection, \OURS{} significantly reduces validation labeling costs without compromising accuracy. 
Key results show that, in certain tasks, the required labeling effort is reduced to below 10\% compared to a fully labeled dataset. 
Our findings emphasize the importance of selecting suitable ensemble methods based on model performance, the superiority of uncertainty sampling over random strategies, and the importance of selecting suitable modes to compose the model committee.

\bibliography{anthology,custom_full}
\clearpage
\appendix

\section{Pseudocode of LEMR}
\label{alg:alg1}
The pseudocode of \OURS{} is shown:

\begin{algorithm}
    \SetKwFunction{isOddNumber}{isOddNumber}
    \SetKwInOut{KwIn}{Input}
    \SetKwInOut{KwOut}{Output}
    \KwIn{model set  $\mathcal{M} = \{ m_{k}  \}_{k \in [K] }$, unlabeled validation set $D_{V}= \{x_{i}\}_{i \in [N]}$, budget $B$, iteration budget $b$.}
    \KwOut{rank $r_{p}$ of $\mathcal{M}$. }
    \scalebox{0.8}{\tcp{Initialization} }\\
    \scalebox{0.8}{$ \mathcal{M} \leftarrow \{ m_{k}  \}_{k \in [K] },   \mathcal{M}_{C}^{ (1)} \leftarrow  \mathcal{M},T \leftarrow \lfloor{\frac{B}{b}}\rfloor , $}
    \scalebox{0.8}{$ L_{g} \leftarrow \emptyset , D_{V} \leftarrow \{x_{i}\}_{i \in [N]} $}\\

     \For{$t = 1 $\KwTo $T$}{
     
        \scalebox{0.8}{\tcp{Initialize the set of pseudo-labels}}
        
        \scalebox{0.8}{$L_{p} \leftarrow \emptyset$ }
        \scalebox{0.8}{\tcp{Step I: Pseudo-label decided by model committee}}

         \For{$x_i$ \KwTo $D_{V}$}{
                    \scalebox{0.8}{$\hat{y}^{ (t)}_{i} \leftarrow  g (x_{i}, \mathcal{M}_{C}^{ (t)})$}
                    
                    \scalebox{0.8}{$L_{p} \leftarrow L_{p} + \{ \hat{y}^{ (t)}_{i}\}$ }
        }
        
        \scalebox{0.8}{\tcp{Step II: Active label acquisition}}

         \scalebox{0.8}{$S^{ (t)}  \leftarrow  l (L_{p}, b) $}

         \For{$x_j$ \KwTo $S^{ (t)}$}{
                    \scalebox{0.8}{$y^{ (t)}_{j} \leftarrow  \textbf{ground-truth label}$}
                    
                    \scalebox{0.8}{$L_{g} \leftarrow L_{g} + \{ y^{ (t)}_{j}\}$ }

                    \scalebox{0.8}{$L_{p} \leftarrow L_{p} - \{ \hat{y}^{ (t)}_{i} \}$ }

                    \scalebox{0.8}{$D_{V} \leftarrow D_{V} - \{ x_{j} \}$ }
        }

        \scalebox{0.8}{\tcp{Step III: Model committee selection}}
        
        \scalebox{0.8}{$\mathcal{M}_{C}^{ (t+1)} \leftarrow s (L_p, L_g,\mathcal{M}) $}

    }

    \scalebox{0.8}{\tcp{Step-IV: Model Ranking}}

    \scalebox{0.8}{$r_{p} \leftarrow r (L_p,L_g , \mathcal{M})$}

    \KwRet{$r_{p}$ }
    \caption{ \OURS{}}
    % \label{alg:alg1}
\end{algorithm}

\begin{table*}[ht]
\centering
\scalebox{0.7}{

\begin{tabular}{@{}c|c|cc|c|c@{}}
\toprule
\bottomrule
\textbf{Training Setting}                           & \textbf{Task Type}                                        & \multicolumn{2}{c|}{\textbf{Dataset}}                               & \textbf{Model/Prompt Number} & \textbf{\# Data} \\  \toprule \toprule
\multirow{8}{*}{\textbf{Weak Supervision}}          & \multirow{2}{*}{\textbf{Sentiment Classification}}   & \multicolumn{2}{c|}{\textbf{Yelp}}                                  & 480                   & 3800             \\ \cmidrule(l){3-6} 
                                                    &                                                      & \multicolumn{2}{c|}{\textbf{IMDB}}                                  & 480                   & 2500             \\ \cmidrule(l){2-6} 
                                                    & \multirow{2}{*}{\textbf{Spam Classification}}        & \multicolumn{2}{c|}{\textbf{SMS}}                                   & 480                   & 500              \\ \cmidrule(l){3-6} 
                                                    &                                                      & \multicolumn{2}{c|}{\textbf{IMDB}}                                  & 480                   & 2500             \\ \cmidrule(l){2-6} 
                                                    & \textbf{Topic Classification}                        & \multicolumn{2}{c|}{\textbf{AGNews}}                                & 480                   & 12000            \\ \cmidrule(l){2-6} 
                                                    & \textbf{Question Classification}                     & \multicolumn{2}{c|}{\textbf{Trec}}                                  & 480                    & 500              \\ \midrule \toprule
\multirow{13}{*}{\textbf{Semi-supervised Learning}} & \multirow{8}{*}{\textbf{Sentiment Classification}}   & \multicolumn{1}{c|}{\multirow{2}{*}{\textbf{IMDB}}}          & 20   & 400                   & 2000             \\ \cmidrule(l){4-6} 
                                                    &                                                      & \multicolumn{1}{c|}{}                                        & 100  & 400                   & 2000             \\ \cmidrule(l){3-6} 
                                                    &                                                      & \multicolumn{1}{c|}{\multirow{2}{*}{\textbf{Yelp Review}}}   & 250  & 400                   & 25000            \\ \cmidrule(l){4-6} 
                                                    &                                                      & \multicolumn{1}{c|}{}                                        & 1000 & 400                   & 25000            \\ \cmidrule(l){3-6} 
                                                    &                                                      & \multicolumn{1}{c|}{\multirow{2}{*}{\textbf{Amazon Review}}} & 250  & 400                   & 25000            \\ \cmidrule(l){4-6} 
                                                    &                                                      & \multicolumn{1}{c|}{}                                        & 1000 & 400                   & 25000            \\ \cmidrule(l){2-6} 
                                                    & \multirow{5}{*}{\textbf{Topic Classification}}       & \multicolumn{1}{c|}{\multirow{2}{*}{\textbf{Yahoo! Answer}}} & 500  & 400                   & 50000            \\ \cmidrule(l){4-6} 
                                                    &                                                      & \multicolumn{1}{c|}{}                                        & 2000 & 400                   & 50000            \\ \cmidrule(l){3-6} 
                                                    &                                                      & \multicolumn{1}{c|}{\multirow{2}{*}{\textbf{AGNews}}}       & 40   & 400                   & 10000            \\ \cmidrule(l){4-6} 
                                                    &                                                      & \multicolumn{1}{c|}{}                                        & 200  & 400                   & 10000            \\ \midrule \toprule
\multirow{11}{*}{\textbf{Prompt Selection}}          & \textbf{Coreference Resolution}                      & \multicolumn{2}{c|}{\textbf{WSC}}                                   & 10                    & 104              \\ \cmidrule(l){2-6} 
                                                    & \textbf{Word Sense Disambiguation}                   & \multicolumn{2}{c|}{\textbf{WiC}}                                   & 10                    & 638              \\ \cmidrule(l){2-6} 
                                                    & \textbf{Sentence Completion}                         & \multicolumn{2}{c|}{\textbf{Story}}                                 & 6                     & 3742             \\ \cmidrule(l){2-6} 
                                                    & \multirow{6}{*}{\textbf{Natural Language Inference}} & \multicolumn{2}{c|}{\textbf{CB}}                                    & 15                    & 56               \\ \cmidrule(l){3-6} 
                                                    &                                                      & \multicolumn{2}{c|}{\textbf{RTE}}                                   & 10                    & 277              \\ \cmidrule(l){3-6} 
                                                    &                                                      & \multicolumn{2}{c|}{\textbf{ANLI1}}                                 & 15                    & 1000             \\ \cmidrule(l){3-6} 
                                                    &                                                      & \multicolumn{2}{c|}{\textbf{ANLI2}}                                 & 15                    & 1000             \\ \cmidrule(l){3-6} 
                                                    &                                                      & \multicolumn{2}{c|}{\textbf{ANLI3}}                                 & 15                    & 1200             \\ \bottomrule \bottomrule
\end{tabular}

}
\caption{The initial model set included in MoraBench and the total size of the validation set plus the test set, i.e., \textbf{\# Data}. 
The number after the dataset of Semi-supervised Learning indicates the number of labels used in semi-supervised training stage.
The description of datasets and generation configuration for each model set are given in the Appendix~\ref{sec:dataset} and Appendix~\ref{sec:modelset-gen}.
We plan to add more model set soon.
}
\label{tab:dataset_sum}
\end{table*}

\begin{table*}[t]
\centering
\scalebox{0.60}{
\begin{tabular}{@{}lllccccll@{}}
\toprule
\bottomrule
\multicolumn{3}{c|}{\textbf{Method}}                                                                                                                    & \multicolumn{5}{c}{\textbf{Dataset}}                                                                                          &                           \\ \midrule
\multicolumn{1}{c|}{\begin{tabular}[c]{@{}c@{}}\textbf{Pseudo-label}\\ \textbf{Generation}\end{tabular}}                                & \multicolumn{1}{c|}{\begin{tabular}[c]{@{}c@{}}\textbf{Active Label}\\ \textbf{Acquisition}\end{tabular}}                                 & \multicolumn{1}{c|}{\textbf{\begin{tabular}[c]{@{}c@{}}\textbf{Model Committee}\\ \textbf{Selection}\end{tabular}}}          & \textbf{Yelp}                 & \textbf{SMS}                  & \textbf{IMDB}                 & \textbf{AGNews}               & \multicolumn{1}{c|}{\textbf{Trec}} & \multicolumn{1}{c}{\textbf{Avg.}}  \\ \midrule
\multicolumn{1}{l|}{\multirow{10}{*}{Hard Ensemble}} & \multicolumn{1}{l|}{\multirow{2}{*}{Random}}          & \multicolumn{1}{l|}{All-model} & \multicolumn{1}{c}{522} & \multicolumn{1}{c}{88} & \multicolumn{1}{c}{482} & \multicolumn{1}{c}{1463} & \multicolumn{1}{c}{99} & \multicolumn{1}{|c}{530.8} \\ \cmidrule (lr){3-3}
\multicolumn{1}{l|}{}                                & \multicolumn{1}{l|}{}                                 & \multicolumn{1}{l|}{Z-score}   & \multicolumn{1}{c}{522} & \multicolumn{1}{c}{88} & \multicolumn{1}{c}{482} & \multicolumn{1}{c}{1463} & \multicolumn{1}{c}{99} & \multicolumn{1}{|c}{530.8} \\ \cmidrule (lr){2-3}
\multicolumn{1}{l|}{}                                & \multicolumn{1}{l|}{\multirow{2}{*}{Uncertainty}}     & \multicolumn{1}{l|}{All-model} & \multicolumn{1}{c}{378} & \multicolumn{1}{c}{7} & \multicolumn{1}{c}{386} & \multicolumn{1}{c}{210} & \multicolumn{1}{c}{59} & \multicolumn{1}{|c}{208.0} \\ \cmidrule (lr){3-3}
\multicolumn{1}{l|}{}                                & \multicolumn{1}{l|}{}                                 & \multicolumn{1}{l|}{Z-score}   & \multicolumn{1}{c}{340} & \multicolumn{1}{c}{7} & \multicolumn{1}{c}{295} & \multicolumn{1}{c}{211} & \multicolumn{1}{c}{56} & \multicolumn{1}{|c}{181.8} \\ \cmidrule (lr){2-3}
\multicolumn{1}{l|}{}                                & \multicolumn{1}{l|}{\multirow{2}{*}{Margin}}          & \multicolumn{1}{l|}{All-model} & \multicolumn{1}{c}{378} & \multicolumn{1}{c}{7} & \multicolumn{1}{c}{386} & \multicolumn{1}{c}{211} & \multicolumn{1}{c}{60} & \multicolumn{1}{|c}{208.4} \\ \cmidrule (lr){3-3}
\multicolumn{1}{l|}{}                                & \multicolumn{1}{l|}{}                                 & \multicolumn{1}{l|}{Z-score}   & \multicolumn{1}{c}{340} & \multicolumn{1}{c}{7} & \multicolumn{1}{c}{295} & \multicolumn{1}{c}{211} & \multicolumn{1}{c}{57} & \multicolumn{1}{|c}{182.0} \\  \cmidrule (lr){2-3}
\multicolumn{1}{l|}{}                                & \multicolumn{1}{l|}{\multirow{2}{*}{Entropy}}         & \multicolumn{1}{l|}{All-model} & \multicolumn{1}{c}{378} & \multicolumn{1}{c}{7} & \multicolumn{1}{c}{386} & \multicolumn{1}{c}{220} & \multicolumn{1}{c}{59} & \multicolumn{1}{|c}{210.0} \\ \cmidrule (lr){3-3}
\multicolumn{1}{l|}{}                                & \multicolumn{1}{l|}{}                                 & \multicolumn{1}{l|}{Z-score}   & \multicolumn{1}{c}{340} & \multicolumn{1}{c}{7} & \multicolumn{1}{c}{295} & \multicolumn{1}{c}{215} & \multicolumn{1}{c}{55} & \multicolumn{1}{|c}{182.4} \\ \toprule
\multicolumn{1}{l|}{\multirow{10}{*}{Soft Ensemble}} & \multicolumn{1}{l|}{\multirow{2}{*}{Random}}          & \multicolumn{1}{l|}{All-model} & \multicolumn{1}{c}{529} & \multicolumn{1}{c}{88} & \multicolumn{1}{c}{482} & \multicolumn{1}{c}{1445} & \multicolumn{1}{c}{99} & \multicolumn{1}{|c}{528.6} \\ \cmidrule (lr){3-3}
\multicolumn{1}{l|}{}                                & \multicolumn{1}{l|}{}                                 & \multicolumn{1}{l|}{Z-score}  & \multicolumn{1}{c}{529} & \multicolumn{1}{c}{88} & \multicolumn{1}{c}{482} & \multicolumn{1}{c}{1445} & \multicolumn{1}{c}{99} & \multicolumn{1}{|c}{528.6} \\ \cmidrule (lr){2-3}
\multicolumn{1}{l|}{}                                & \multicolumn{1}{l|}{\multirow{2}{*}{Uncertainty}}     & \multicolumn{1}{l|}{All-model} & \multicolumn{1}{c}{415} & \multicolumn{1}{c}{8} & \multicolumn{1}{c}{283} & \multicolumn{1}{c}{210} & \multicolumn{1}{c}{66} & \multicolumn{1}{|c}{196.4} \\ \cmidrule (lr){3-3}
\multicolumn{1}{l|}{}                                & \multicolumn{1}{l|}{}                                 & \multicolumn{1}{l|}{Z-score}   & \multicolumn{1}{c}{379} & \multicolumn{1}{c}{8} & \multicolumn{1}{c}{297} & \multicolumn{1}{c}{210} & \multicolumn{1}{c}{66} & \multicolumn{1}{|c}{192.0} \\ \cmidrule (lr){2-3}
\multicolumn{1}{l|}{}                                & \multicolumn{1}{l|}{\multirow{2}{*}{Margin}}          & \multicolumn{1}{l|}{All-model} & \multicolumn{1}{c}{415} & \multicolumn{1}{c}{8} & \multicolumn{1}{c}{283} & \multicolumn{1}{c}{219} & \multicolumn{1}{c}{66} & \multicolumn{1}{|c}{198.2} \\ \cmidrule (lr){3-3}
\multicolumn{1}{l|}{}                                & \multicolumn{1}{l|}{}                                 & \multicolumn{1}{l|}{Z-score}   & \multicolumn{1}{c}{379} & \multicolumn{1}{c}{8} & \multicolumn{1}{c}{297} & \multicolumn{1}{c}{220} & \multicolumn{1}{c}{66} & \multicolumn{1}{|c}{194.0} \\ \cmidrule (lr){2-3}
\multicolumn{1}{l|}{}                                & \multicolumn{1}{l|}{\multirow{2}{*}{Entropy}}         & \multicolumn{1}{l|}{All-model} & \multicolumn{1}{c}{415} & \multicolumn{1}{c}{8} & \multicolumn{1}{c}{283} & \multicolumn{1}{c}{219} & \multicolumn{1}{c}{67} & \multicolumn{1}{|c}{198.4} \\ \cmidrule (lr){3-3}
\multicolumn{1}{l|}{}                                & \multicolumn{1}{l|}{}                                 & \multicolumn{1}{l|}{Z-score}   & \multicolumn{1}{c}{379} & \multicolumn{1}{c}{8} & \multicolumn{1}{c}{297} & \multicolumn{1}{c}{214} & \multicolumn{1}{c}{66} & \multicolumn{1}{|c}{192.8} \\
\toprule
\multicolumn{3}{c|}{\textbf{Size of Unlabeled Validation Set} $D_{V}$} & \multicolumn{1}{c}{760} & \multicolumn{1}{c}{100} & \multicolumn{1}{c}{500} & \multicolumn{1}{c}{2400} & \multicolumn{1}{c}{100} & \multicolumn{1}{|c}{-} \\ 

\bottomrule
\bottomrule
\end{tabular}

}
\caption{\textbf{Weak supervision setting}: 
This table illustrates the minimum labeling budget necessary to achieve an optimal gap of zero in our framework.
}
\label{tab:wrench_optimal_gap_equal_one}
\end{table*}

\begin{table*}[t]
\centering
\scalebox{0.60}{

\begin{tabular}{@{}llllllllllll@{}}
\toprule
\bottomrule
\multicolumn{3}{c|}{\textbf{Method}}                                                                                                                                                                                                                                                                  & \multicolumn{8}{c}{\textbf{Dataset}}                                                                                                                                                                                           &                           \\ \midrule
\multicolumn{1}{c}{\textbf{\begin{tabular}[c]{@{}c@{}}Pseudo-label\\ Generation\end{tabular}}} & \multicolumn{1}{c}{\textbf{\begin{tabular}[c]{@{}c@{}}Active Label\\ Acquisition\end{tabular}}} & \multicolumn{1}{c|}{\textbf{\begin{tabular}[c]{@{}c@{}}Model Committee\\ Selection\end{tabular}}} & \multicolumn{1}{c}{\textbf{WSC}}   & \multicolumn{1}{c}{\textbf{Story}} & \multicolumn{1}{c}{\textbf{CB}}    & \multicolumn{1}{c}{\textbf{RTE}}   & \multicolumn{1}{c}{\textbf{WiC}}   & \multicolumn{1}{c}{\textbf{ANLI1}} & \multicolumn{1}{c}{\textbf{ANLI2}} & \multicolumn{1}{c|}{\textbf{ANLI3}} & \multicolumn{1}{c}{\textbf{Avg.}}  \\ \midrule

\multicolumn{1}{l|}{\multirow{10}{*}{Hard Ensemble}}                                            & \multicolumn{1}{l|}{\multirow{2}{*}{Random}}                                                    & \multicolumn{1}{l|}{All Model}                                                                    & \multicolumn{1}{c}{19} & \multicolumn{1}{c}{216} & \multicolumn{1}{c}{10} & \multicolumn{1}{c}{11} & \multicolumn{1}{c}{102} & \multicolumn{1}{c}{44} & \multicolumn{1}{c}{194} & \multicolumn{1}{c|}{237} & \multicolumn{1}{c}{105.03} \\   \cmidrule(lr){3-3}
\multicolumn{1}{l|}{}                                                                          & \multicolumn{1}{l|}{}                                                                           & \multicolumn{1}{l|}{Z-score}                                                                      & \multicolumn{1}{c}{19} & \multicolumn{1}{c}{216} & \multicolumn{1}{c}{10} & \multicolumn{1}{c}{11} & \multicolumn{1}{c}{102} & \multicolumn{1}{c}{44} & \multicolumn{1}{c}{194} & \multicolumn{1}{c|}{237} & \multicolumn{1}{c}{105.03} \\   \cmidrule(lr){2-3}
\multicolumn{1}{l|}{}                                                                          & \multicolumn{1}{l|}{\multirow{2}{*}{Uncertainty}}                                               & \multicolumn{1}{l|}{All Model}                                                                    & \multicolumn{1}{c}{3} & \multicolumn{1}{c}{216} & \multicolumn{1}{c}{5} & \multicolumn{1}{c}{4} & \multicolumn{1}{c}{36} & \multicolumn{1}{c}{20} & \multicolumn{1}{c}{166} & \multicolumn{1}{c|}{201} & \multicolumn{1}{c}{81.79} \\   \cmidrule(lr){3-3}
\multicolumn{1}{l|}{}                                                                          & \multicolumn{1}{l|}{}                                                                           & \multicolumn{1}{l|}{Z-score}                                                                      & \multicolumn{1}{c}{1} & \multicolumn{1}{c}{44} & \multicolumn{1}{c}{1} & \multicolumn{1}{c}{4} & \multicolumn{1}{c}{43} & \multicolumn{1}{c}{12} & \multicolumn{1}{c}{192} & \multicolumn{1}{c|}{232} & \multicolumn{1}{c}{67.05} \\   \cmidrule(lr){2-3}
\multicolumn{1}{l|}{}                                                                          & \multicolumn{1}{l|}{\multirow{2}{*}{Margin}}                                                    & \multicolumn{1}{l|}{All Model}                                                                    & \multicolumn{1}{c}{3} & \multicolumn{1}{c}{216} & \multicolumn{1}{c}{4} & \multicolumn{1}{c}{4} & \multicolumn{1}{c}{36} & \multicolumn{1}{c}{6} & \multicolumn{1}{c}{166} & \multicolumn{1}{c|}{201} & \multicolumn{1}{c}{79.84} \\   \cmidrule(lr){3-3}
\multicolumn{1}{l|}{}                                                                          & \multicolumn{1}{l|}{}                                                                           & \multicolumn{1}{l|}{Z-score}                                                                      & \multicolumn{1}{c}{1} & \multicolumn{1}{c}{44} & \multicolumn{1}{c}{1} & \multicolumn{1}{c}{4} & \multicolumn{1}{c}{43} & \multicolumn{1}{c}{18} & \multicolumn{1}{c}{192} & \multicolumn{1}{c|}{232} & \multicolumn{1}{c}{67.74} \\   \cmidrule(lr){2-3}
\multicolumn{1}{l|}{}                                                                          & \multicolumn{1}{l|}{\multirow{2}{*}{Entropy}}                                                   & \multicolumn{1}{l|}{All Model}                                                                    & \multicolumn{1}{c}{3} & \multicolumn{1}{c}{216} & \multicolumn{1}{c}{5} & \multicolumn{1}{c}{4} & \multicolumn{1}{c}{36} & \multicolumn{1}{c}{20} & \multicolumn{1}{c}{168} & \multicolumn{1}{c|}{206} & \multicolumn{1}{c}{82.60} \\   \cmidrule(lr){3-3}
\multicolumn{1}{l|}{}                                                                          & \multicolumn{1}{l|}{}                                                                           & \multicolumn{1}{l|}{Z-score}                                                                      & \multicolumn{1}{c}{1} & \multicolumn{1}{c}{44} & \multicolumn{1}{c}{1} & \multicolumn{1}{c}{4} & \multicolumn{1}{c}{43} & \multicolumn{1}{c}{22} & \multicolumn{1}{c}{194} & \multicolumn{1}{c|}{228} & \multicolumn{1}{c}{67.66} \\   \toprule
\multicolumn{1}{l|}{\multirow{10}{*}{Soft Ensemble}}                                            & \multicolumn{1}{l|}{\multirow{2}{*}{Random}}                                                    & \multicolumn{1}{l|}{All Model}                                                                    & \multicolumn{1}{c}{18} & \multicolumn{1}{c}{748} & \multicolumn{1}{c}{10} & \multicolumn{1}{c}{52} & \multicolumn{1}{c}{30} & \multicolumn{1}{c}{57} & \multicolumn{1}{c}{194} & \multicolumn{1}{c|}{237} & \multicolumn{1}{c}{168.20} \\   \cmidrule(lr){3-3}
\multicolumn{1}{l|}{}                                                                          & \multicolumn{1}{l|}{}                                                                           & \multicolumn{1}{l|}{Z-score}                                                                      & \multicolumn{1}{c}{18} & \multicolumn{1}{c}{748} & \multicolumn{1}{c}{10} & \multicolumn{1}{c}{52} & \multicolumn{1}{c}{30} & \multicolumn{1}{c}{57} & \multicolumn{1}{c}{194} & \multicolumn{1}{c|}{237} & \multicolumn{1}{c}{168.20} \\   \cmidrule(lr){2-3}
\multicolumn{1}{l|}{}                                                                          & \multicolumn{1}{l|}{\multirow{2}{*}{Uncertainty}}                                               & \multicolumn{1}{l|}{All Model}                                                                    & \multicolumn{1}{c}{5} & \multicolumn{1}{c}{142} & \multicolumn{1}{c}{6} & \multicolumn{1}{c}{32} & \multicolumn{1}{c}{43} & \multicolumn{1}{c}{184} & \multicolumn{1}{c}{184} & \multicolumn{1}{c|}{225} & \multicolumn{1}{c}{102.54} \\  \cmidrule(lr){3-3}
\multicolumn{1}{l|}{}                                                                          & \multicolumn{1}{l|}{}                                                                           & \multicolumn{1}{l|}{Z-score}                                                                      & \multicolumn{1}{c}{2} & \multicolumn{1}{c}{59} & \multicolumn{1}{c}{1} & \multicolumn{1}{c}{4} & \multicolumn{1}{c}{50} & \multicolumn{1}{c}{194} & \multicolumn{1}{c}{188} & \multicolumn{1}{c|}{225} & \multicolumn{1}{c}{90.67} \\   \cmidrule(lr){2-3}
\multicolumn{1}{l|}{}                                                                          & \multicolumn{1}{l|}{\multirow{2}{*}{Margin}}                                                    & \multicolumn{1}{l|}{All Model}                                                                    & \multicolumn{1}{c}{5} & \multicolumn{1}{c}{142} & \multicolumn{1}{c}{3} & \multicolumn{1}{c}{32} & \multicolumn{1}{c}{43} & \multicolumn{1}{c}{184} & \multicolumn{1}{c}{186} & \multicolumn{1}{c|}{228} & \multicolumn{1}{c}{102.34} \\   \cmidrule(lr){3-3}
\multicolumn{1}{l|}{}                                                                          & \multicolumn{1}{l|}{}                                                                           & \multicolumn{1}{l|}{Z-score}                                                                      & \multicolumn{1}{c}{2} & \multicolumn{1}{c}{59} & \multicolumn{1}{c}{1} & \multicolumn{1}{c}{4} & \multicolumn{1}{c}{50} & \multicolumn{1}{c}{194} & \multicolumn{1}{c}{186} & \multicolumn{1}{c|}{225} & \multicolumn{1}{c}{90.54} \\   \cmidrule(lr){2-3}
\multicolumn{1}{l|}{}                                                                          & \multicolumn{1}{l|}{\multirow{2}{*}{Entropy}}                                                   & \multicolumn{1}{l|}{All Model}                                                                    & \multicolumn{1}{c}{5} & \multicolumn{1}{c}{142} & \multicolumn{1}{c}{6} & \multicolumn{1}{c}{32} & \multicolumn{1}{c}{43} & \multicolumn{1}{c}{12} & \multicolumn{1}{c}{184} & \multicolumn{1}{c|}{225} & \multicolumn{1}{c}{81.06} \\   \cmidrule(lr){3-3}
\multicolumn{1}{l|}{}                                                                          & \multicolumn{1}{l|}{}                                                                           & \multicolumn{1}{l|}{Z-score}                                                                      & \multicolumn{1}{c}{2} & \multicolumn{1}{c}{59} & \multicolumn{1}{c}{1} & \multicolumn{1}{c}{4} & \multicolumn{1}{c}{50} & \multicolumn{1}{c}{12} & \multicolumn{1}{c}{188} & \multicolumn{1}{c|}{225} & \multicolumn{1}{c}{67.83} \\

\toprule
\multicolumn{3}{c|}{\textbf{Size of Unlabeled Validation Set} $D_{V}$} & \multicolumn{1}{c}{20} & \multicolumn{1}{c}{748} & \multicolumn{1}{c}{11} & \multicolumn{1}{c}{55} & \multicolumn{1}{c}{127} & \multicolumn{1}{c}{200} & \multicolumn{1}{c}{200} & \multicolumn{1}{c}{240} & \multicolumn{1}{|c}{-} \\  

\bottomrule
\bottomrule

\end{tabular}

}
\caption{
\textbf{Prompt selection setting}: 
This table illustrates the minimum labeling budget necessary to achieve an optimal gap of zero in our framework.
}
\label{tab:prompt_optimal_gap}
\end{table*}

\section{Experiments Setup}
\label{sec:exsetup}
Here, we show the implementation details and design space of our paper.

\subsection{Implementation Details}
Our experimental environment is configured on a high-performance computing setup, comprising an Intel (R) Xeon (R) Platinum 8358P CPU clocked at 2.60GHz, backed by a substantial 512GB of memory. The computational muscle is provided by eight NVIDIA A40 GPUs, each with a hefty 48GB of memory. 
For model set generation (detailed in Appendix~\ref{sec:modelset-gen}), models are evaluated on validation and test datasets at regular intervals during training, with all outputs saved. 
These outputs are then divided using a 2:8 ratio to create validation and test sets for model selection. This process is repeated across 50 different splits, and the resulting data is averaged, ensuring a reliable and consistent foundation for our model selection analysis.

\subsection{Design Space}
Based on Step-I  (Section~\ref{sec:plg}), Step-II  (Section~\ref{sec:tlr}) and Step-III  (Section~\ref{sec:msu}), our design space $\mathcal{D}$ can be defined as:
\begin{equation}
\resizebox{0.8\linewidth}{!}{
  $\begin{aligned}
  \mathcal{D} &= \textbf{\{Hard ensemble, Soft ensemble\}}\\
  &\times \textbf{\{ Uncertainty, Margin, Entropy, Random\}}\\
  &\times  \textbf{\{ Z-score, All-model\}}.
  \end{aligned}$
}
\end{equation}

Therefore, there will be a total of $2  \times  4 \times  2 = 20$ method combinations within our framework.

\section{Model Set Generation Setups}
\label{sec:modelset-gen}
The statistics of all model sets within MoraBench are shown in Table~\ref{tab:dataset_sum}.

\subsection{Generation Setups for Semi-supervised Learning Setting}
\label{sec:semi-setting}
Leveraging the USB benchmark\footnote{\url{http://github.com/microsoft/Semi-supervised-learning}}~\citet{wang2022usb}, model outputs were obtained from 12 semi-supervised methods across five datasets: IMDB~\cite{IMDB}, Amazon Review~\cite{mcauley2013hidden}, Yelp Review~\cite{yelpwebsite}, AGNews~\cite{AGNews} and Yahoo! Answer~\cite{chang2008importance}. 
More details of these datasets are provided in Appendix~\ref{sec:MSdataset}.

Specially, we use 14 common semi-supervised methods: $\Pi$ model~\cite{rasmus2015semi}, Pseudo Labeling~\cite{lee2013pseudo}, Mean Teacher~\cite{tarvainen2017mean}, VAT~\cite{miyato2018virtual}, MixMatch~\cite{berthelot2019mixmatch}, ReMixMatch~\cite{berthelot2019remixmatch}, UDA~\cite{xie2020unsupervised}, FixMatch~\cite{sohn2020fixmatch}, Dash~\cite{xu2021dash}, CoMatch~\cite{li2021comatch}, CRMatch~\cite{fan2021revisiting}, FlexMatch~\cite{zhang2021flexmatch}, AdaMatch~\cite{berthelot2021adamatch} and SimMatch~\cite{zheng2022simmatch} to generate our model sets in semi-supervised learning setting with dataset we mentioned above.
For detailed training configurations, refer to this website\footnote{\url{http://github.com/microsoft/Semi-supervised-learning/tree/main/config/usb_nlp}}.
We save the model's output every 256 steps. 
Eventually, each method will get 400 outputs.
This means that for each dataset we will have $400 \times 14  = 5600$ model outputs.
In this paper, we randomly selected 10\% of the models from each dataset for model selection.

\subsection{Generation Setups for Weak Supervision Setting}
\label{sec:weak-setting}

Utilizing the WRENCH\footnote{\url{http://github.com/JieyuZ2/wrench}}~\cite{zhang2021wrench} framework, we generated model outputs within a weak supervision setting. 
We generate model outputs across 48 distinct weak supervision configurations on five datasets: SMS~\cite{sms}, AGNews~\cite{AGNews}, Yelp~\cite{AGNews}, IMDB~\cite{IMDB}, Trec~\cite{Li2002LearningQC}.
Specifics on datasets are in Appendix~\ref{sec:MSdataset}.

Specifically,  we follow the training configuration of WRENCH  for model training for the model set, involving an array of label models, label types, model backbones, and varied learning rates.

\textbf{Label Models}: Incorporating Snorkel~\cite{ratner2017snorkel}, majority voting, weighted majority voting~\cite{penrose1946elementary}, and generative model~\cite{bach2017learning}, each offering unique approaches to producing weak labels.

\textbf{Label Types}: Utilization of both soft and hard labels for pseudo-label generation.

\textbf{Model Backbones}: Adoption of bert-base and roberta-base backbones, known for their efficacy in natural language processing.

\textbf{Learning Rates}: Training across three learning rates ($10^{-1}$, $10^{-3}$, and $10^{-5}$) to generate model for model set.

For detailed configuration, refer to the WRENCH repository\footnote{\url{http://github.com/JieyuZ2/wrench/tree/main}}. 
This setup aims to test model selection methods extensively by leveraging a comprehensive and diverse approach to model generation.

\subsection{Generation Setups for Prompt Selection Setting}
\label{sec:prompt-setting}
We employed large language models like GPT-4~\cite{OpenAI2023GPT4TR} and various prompts to generate diverse outputs, assessed using the T0 benchmark\footnote{\url{http://github.com/bigscience-workshop/T0}}~\cite{sanh2021multitask}. This process covered eight tasks, with further information in Appendix~\ref{sec:PSdataset}.
In particular, we adopt the T0 benchmark with eight different datasets. 
The prompts we use for prompt selection all come from the promptsource\footnote{\url{http://github.com/bigscience-workshop/promptsource}}.

\section{Optimal Gap with Different Budget Ratio}
\begin{figure*}[ht]
\centering
\includegraphics[scale=0.192]{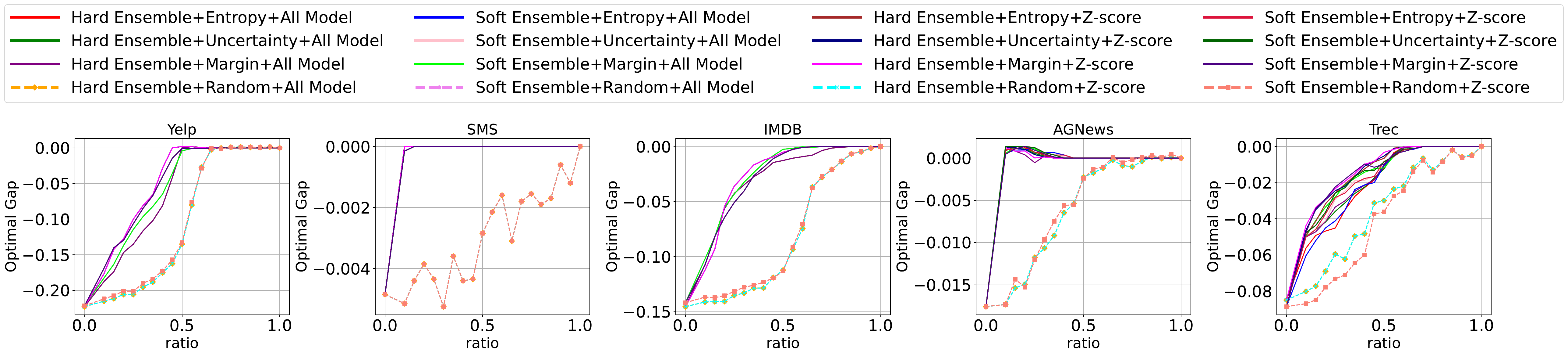}
\caption{\textbf{Weak supervision setting}: This figure illustrates the changes in optimal gap values within our design space.
These changes are observed across budget ratios from 0 to 1.}
\label{fig:wrench_og}
\end{figure*}

\begin{figure*}[ht]
\centering
\includegraphics[scale=0.192]{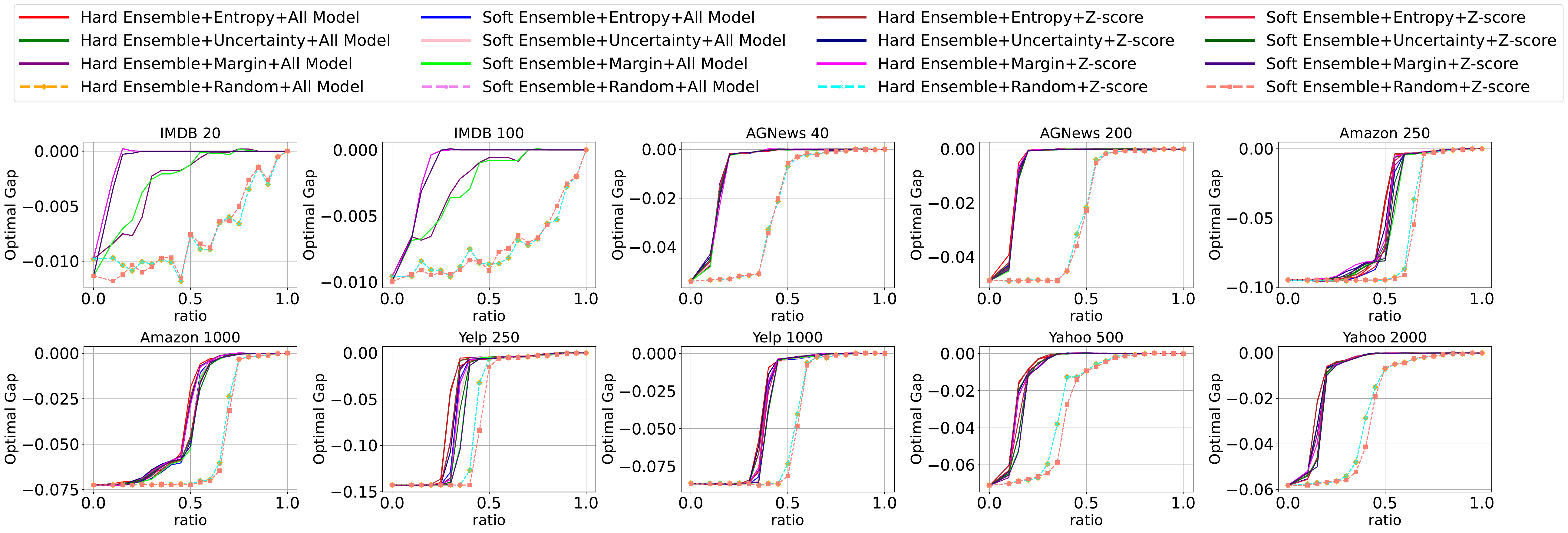}
\caption{\textbf{Semi-supervised learning setting}:
This figure illustrates the changes in optimal gap values within our design space, under a semi-supervised learning setting. 
These changes are observed across budget ratios from 0 to 1.}
\label{fig:usb_og}
\end{figure*}

\begin{figure*}[ht]
\centering
\includegraphics[scale=0.195]{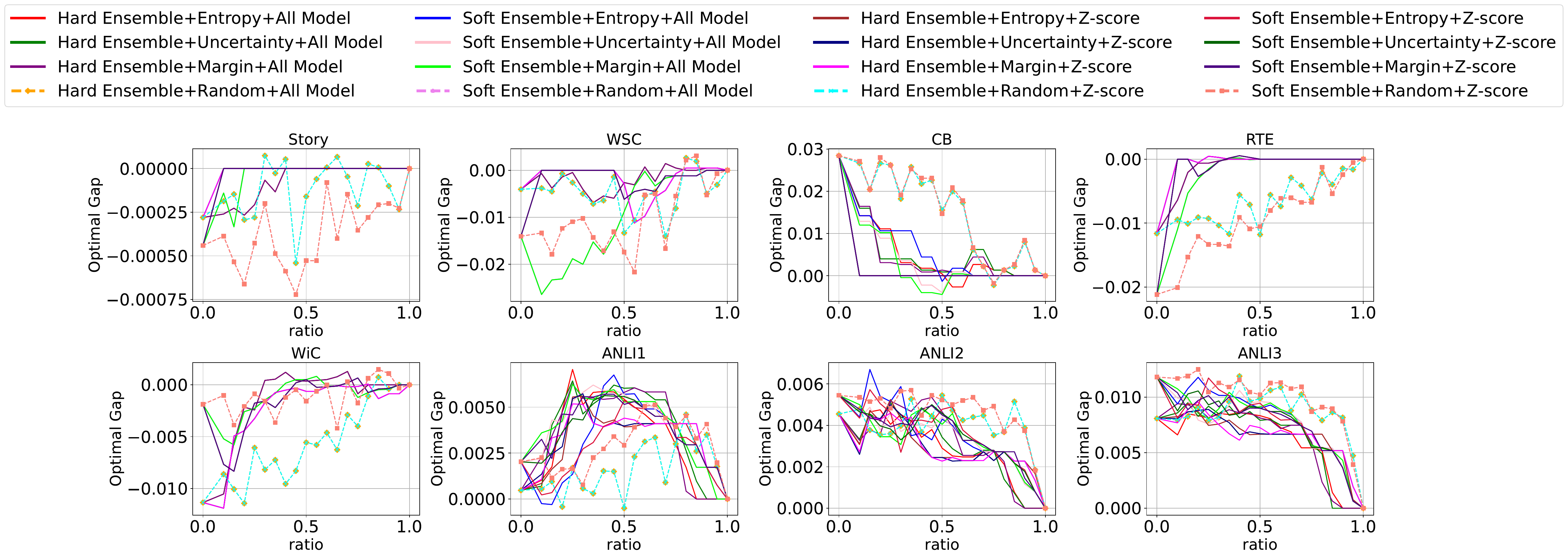}
\caption{\textbf{Prompt selection setting}:
This figure illustrates the changes in optimal gap values within our design space. 
These changes are observed across budget ratios from 0 to 1.}
\label{fig:prompt_og}
\end{figure*}

Our analysis, illustrated in Figures~\ref{fig:wrench_og}, \ref{fig:usb_og}, and \ref{fig:prompt_og}, explores the optimal gap in varying budget ratios, which span from 0 to 1. 
This investigation across diverse scenarios establishes a key insight: 
the existing practice of fully labeling the validation set is wasteful, and we do not need to label the entire validation set in the process of model selection.
This finding further demonstrates the value of \OURS{}, highlighting its ability to optimize resource utilization while maintaining high model selection performance.

\section{Datasets Details}
\label{sec:dataset}

\subsection{Model Selection Datasets}
\label{sec:MSdataset}
\paragraph{SMS~\cite{sms}}. 
This dataset contains 4,571 text messages labeled as spam/not-spam, out of which 500 are held out for validation and 2719 for testing. The labeling functions are generated manually by \cite{Awasthi2020Learning}, including 16 keyword-based and 57 regular expression-based rules.

\paragraph{AGNews~\cite{AGNews}}.
This dataset is a collection of more than one million news articles. It is constructed by \cite{ren2020denoising} choosing the 4 largest topic classes from the original corpus. The total number of training samples is 96K and both validation and testing are 12K. The labeling functions are also generated by \cite{ren2020denoising}, including 9 keyword-based rules.

\paragraph{Yelp~\cite{AGNews}}.
This dataset is a subset of Yelp's businesses, reviews, and user data for binary sentiment classification. It is constructed by \cite{ren2020denoising}, including 30.4K training samples, 3.8K validation samples, and 3.8K testing samples. The labeling functions are also generated by \cite{ren2020denoising}, including 7 heuristic rules on keywords and 1 third-party model on polarity of sentiment.

\paragraph{IMDB~\cite{IMDB}}.
This is a dataset for binary sentiment classification containing a set of 20,000 highly polar movie reviews for training, 2,500 for validation and 2,500 for testing. It is constructed by \cite{ren2020denoising}. The labeling functions are also generated by \cite{ren2020denoising}, including 4 heuristic rules on keywords and 1 heuristic rules on expressions.

\paragraph{Amazon Review~\cite{mcauley2013hidden}.}
This dataset is a sentiment classification dataset. There are 5 classes  (scores). Each class  (score) contains 600,000 training samples and 130,000 test samples. For USB, we draw 50,000 samples and 5,000 samples per class from training samples to form the training dataset and validation dataset respectively. The test dataset is unchanged.

\paragraph{Yelp Review~\cite{yelpwebsite}}
This sentiment classification dataset has 5 classes  (scores). Each class  (score) contains 130,000 training samples and 10,000 test samples. For USB, we draw 50,000 samples and 5,000 samples per class from training samples to form the training dataset and validation dataset respectively. The test dataset is unchanged.

\paragraph{Trec~\cite{Li2002LearningQC}}.
This dataset contains 4,965 labeled questions in the training set, 500 for the validation set, and another 500 for the testing set. It has 6 classes. The labeling functions are generated by \cite{Awasthi2020Learning}, including 68 keyword-based rules.

\paragraph{Yahoo! Answer~\cite{chang2008importance}.}
This dataset has 10 categories. Each class contains 140,000 training samples and 6,000 test samples. For USB, we draw 50,000 samples and 5,000 samples per class from training samples to form the training dataset and validation dataset respectively. The test dataset is unchanged.

\subsection{Prompt Selection Datasets}
\label{sec:PSdataset}
We follow the T0 benchmark~\citep{sanh2021multitask}.
Specifically, the test tasks include natural language inference (RTE~\citep{2005_RTE}, CB~\citep{de2019_CB}, ANLI/R1-R3~\citep{NieWDBWK20_ANLI}), sentence completion (StoryCloze~\citep{story_cloze}), word sense disambiguation (WiC~\citep{wic-paper}), and coreference resolution (WSC~\citep{WSC2012}).

\section{Minimum  Budget to achieve an
optimal gap of zero in other Cases}

We further explored the minimal budget necessary to achieve a zero optimal gap in  weak supervision and  prompt selection setting, with findings presented in Table~\ref{tab:wrench_optimal_gap_equal_one} and Table~\ref{tab:prompt_optimal_gap}.
We can conclude consistent with the text.

To be specific, Firstly, our framework, combined with an appropriate selection of methods, significantly lowers the labeling cost for validation sets.
As seen in \ref{tab:wrench_optimal_gap_equal_one}, for the AGNews task, where only 210 samples need labeling as opposed to labeling 2400 samples of the entire validation set.
This efficiency is further evidenced in the Story task, where selecting the optimal model entails labeling a mere 44 samples instead of the full 748, as shown in Table~\ref{tab:prompt_optimal_gap}.

Then, we can find uncertainty sampling strategy is much better than random strategy.
This is evident in Table~\ref{tab:wrench_optimal_gap_equal_one} and Table~\ref{tab:prompt_optimal_gap}, where uncertainty sampling consistently requires a smaller budget across all tasks. 

Finally, adopting the Z-score method generally reduces labeling costs.
Table \ref{tab:wrench_optimal_gap_equal_one} demonstrates that the Z-score method requires a lesser budget to select the equivalent model as the All-model approach.
This trend is also evident in Table~\ref{tab:prompt_optimal_gap}, where the Z-score variant requires less budget to achieve an optimal gap of 0 compared to the All-model scenario.

\section{Limitations and Potential Risks}
Our evaluations primarily focus on NLP tasks. Although \OURS{} shows promising results in these areas, its effectiveness and adaptability to other domains, such as computer vision or audio processing, remain to be thoroughly investigated. Different domains may exhibit unique challenges, including higher dimensional data or different notions of uncertainty, which could affect the performance of our proposed methods. Besides, the models selected by frameworks like \OURS{} are often deployed in applications with wide-reaching societal impacts. From enhancing educational tools and healthcare diagnostics to improving environmental monitoring, the potential for positive societal impact is vast. However, careful consideration of the implications of these applications, including ethical, social, and environmental impacts, is essential to ensure that they contribute positively to society.

\end{document}